\documentclass[10pt, a4paper, onecolumn, twoside]{amsart}

\usepackage{titlesec, color, amsthm}
\definecolor{blue1}{rgb}{0,0,0}
\renewcommand\thesection{\arabic{section}}

\titleformat{\section}[hang]{\color{blue1}\large\bfseries\sffamily}{\thesection}{0mm}{. }[]
\titleformat{\subsection}[hang] {\color{blue1}\bfseries\sffamily}{\thesubsection}{0em}{. }[]
\titleformat{\subsubsection}[hang] {\color{blue1}\sffamily}{\thesubsubsection}{0em}{. }[]
\titlespacing*{\section}{1em}{3.5ex plus .2ex minus .2ex}{1ex plus .2ex}
\titlespacing*{\subsection}{0em}{3ex plus .2ex minus .2ex}{1ex plus .2ex}
\titlespacing*{\subsubsection}{0em}{3ex plus .2ex minus .2ex}{1ex plus .2ex}

\renewenvironment{abstract}{{\color{blue1}\small\bfseries Abstract.}\footnotesize}{\par \vskip .1in}
\makeatletter
\def\@setauthors{
\begingroup 
\def \thanks{\protect\thanks@warning}
\trivlist \centering\footnotesize \@topsep30\p@\relax \advance\@topsep by -\baselineskip
\item\relax \author@andify \authors \def\\{\protect\linebreak} {\color{blue1}\large\authors} \endtrivlist \endgroup}
\def\@settitle{\centering{\color{blue1} \Large \bfseries \bfseries \@title \par}}
\makeatother
\usepackage[normal, small, up, textfont=it]{caption}
\usepackage{url}
\usepackage{graphicx}
\usepackage{subfigure} 
\usepackage{algorithm}
\usepackage{algorithmic}
\usepackage{bm}
\usepackage{amsmath}
\usepackage{amssymb}
\usepackage{enumerate}
\usepackage{mathtools}

\title[A Flexible Convolutional Solver with Application to Photorealistic Style Transfer]{A Flexible Convolutional Solver with Application to\\Photorealistic Style Transfer}
\author{Gilles~Puy}
\author{Patrick~P\'{e}rez}
\thanks{Gilles~Puy and Patrick~P\'{e}rez are with Technicolor, 975 Avenue des Champs Blancs, 35576 Cesson-S\'evign\'e, France.}

\def\etal{\emph{et al.}}
\def\ie{\emph{i.e.}}
\def\eg{\emph{e.g.}}

\newcommand{\Graph}{\ensuremath{\set{G}}}

\newcommand{\Lap}{\ensuremath{\ma{L}}}
\newcommand{\Fou}{\ensuremath{\ma{U}}}
\newcommand{\Eig}{\ensuremath{\ma{\Lambda}}}

\newcommand{\fou}{\ensuremath{\vec{u}}}

\newcommand{\eig}{\ensuremath{\lambda}}

\newcommand{\Sig}{\ensuremath{\ma{X}}}

\newcommand{\Rbb}{\ensuremath{\mathbb{R}}}

\renewcommand{\leq}{\ensuremath{\leqslant}}

\def\diag{\ensuremath{\mathrm{diag}}}

\newcommand{\adjoint}{\ensuremath{{\intercal}}}

\newcommand{\norm}[1]{\ensuremath{\left\| #1\right\|}}
\newcommand{\abs}[1]{\ensuremath{\left| #1 \right|}}

\newcommand{\ma}[1]{\ensuremath{\mathsf{#1}}}
\renewcommand{\vec}[1]{\ensuremath{\bm{#1}}}

\newcommand{\set}[1]{\ensuremath{\mathcal{#1}}}

\renewcommand{\th}{\ensuremath{\text{th}}}

\graphicspath{{figs/}}
\DeclareGraphicsExtensions{.jpg,.png}

\begin{document} 

\maketitle

%
\begin{figure}[h]
\centering
\begin{minipage}{0.32\linewidth}
\hspace{\linewidth}
\end{minipage}
\begin{minipage}{0.32\linewidth}
\centering
\includegraphics[width=28mm]{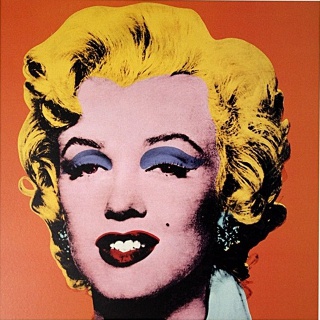}\\
\scriptsize (1) Artistic style
\end{minipage}
\begin{minipage}{0.32\linewidth}
\centering
\includegraphics[width=\linewidth]{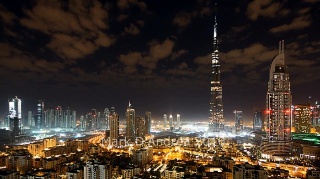}\\
\scriptsize (2) Photorealistic style
\end{minipage}
\\
\begin{minipage}{0.32\linewidth}
\centering
\includegraphics[width=\linewidth]{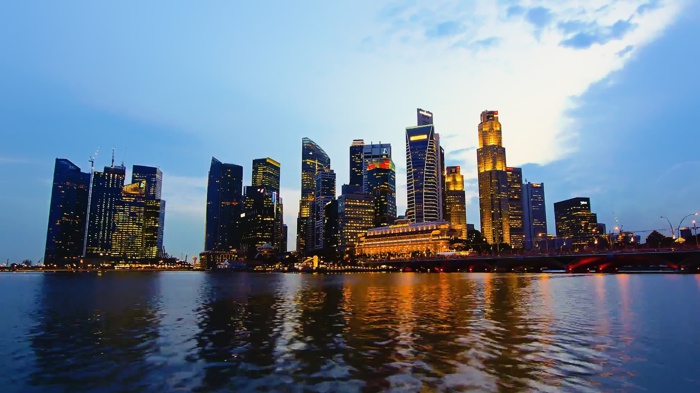}\\
\scriptsize (a) Input image
\end{minipage}
\begin{minipage}{0.32\linewidth}
\centering
\includegraphics[width=\linewidth]{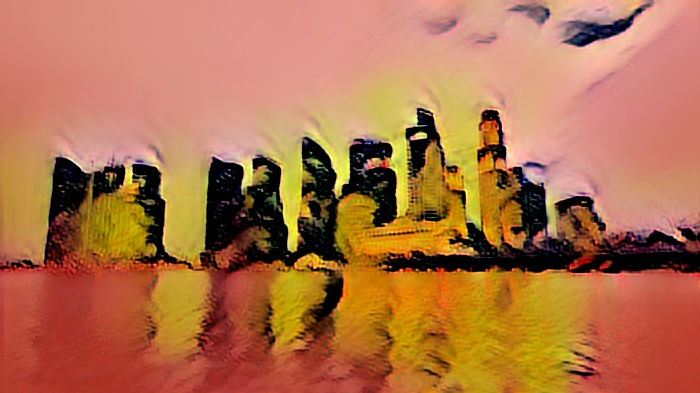}\\
\scriptsize (b) Artistic transfer
\end{minipage}
\begin{minipage}{0.32\linewidth}
\centering
\includegraphics[width=\linewidth]{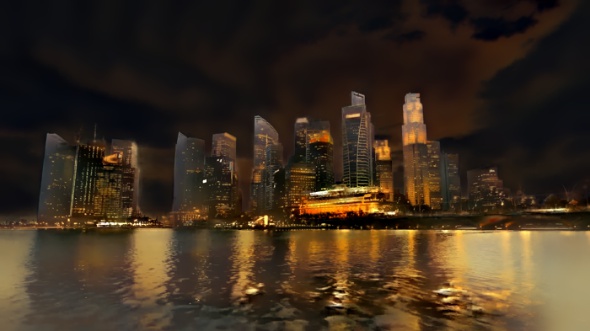}
\scriptsize (c) Photorealistic transfer
\end{minipage}
\\
\caption{Given an input image (a), the proposed convolutional neural network yields fast artistic stylisation results, \eg, (b) for style (1), and can be modified at runtime -- hence \textit{without retraining} -- to yield photorealistic stylisation results, \eg, (c) for style (2).}
\end{figure}
%


%
\begin{abstract}
We propose a new flexible deep convolutional neural network (convnet) to perform fast visual style transfer. In contrast to existing convnets that address the same task, our architecture derives directly from the structure of the gradient descent originally used to solve the style transfer problem \cite{gatys16}. Like existing convnets, ours approximately solves the original problem much faster than the gradient descent. However, our network is uniquely flexible by design: it can be manipulated at runtime to enforce new constraints on the final solution. In particular, we show how to modify it to obtain a photorealistic result with no retraining. We study the modifications made by \cite{luan17} to the original cost function of \cite{gatys16} to achieve photorealistic style transfer. These modifications affect directly the gradient descent and can be reported on-the-fly in our network. These modifications are possible as the proposed architecture stems from unrolling the gradient descent.
\end{abstract}
%

%
\section{Introduction}
Transferring the style of one image onto another one, while keeping the content of the second, is a difficult task but for which visually impressive results have recently been obtained thanks to deep neural networks. 

In \cite{gatys16}, the authors adapt their texture synthesis method \cite{gatys15} to perform style transfer. The stylised image is obtained by minimisation of a loss function built using feature maps obtained from a pre-trained deep neural network. While yielding good results, this method is nevertheless slow as the stylised image is a solution of a long iterative optimisation problem. This issue has however been overcome by researchers who have proposed to train deep networks that estimate rapidly a solution of the style transfer problem. The first architectures were trained to transform any natural image, given as input to the network, to a fixed style \cite{johnson2016perceptual,ulyanov2016texture}. These networks thus need to be retrained for every new style. This drawback was partly addressed by works that showed that the style given to an image was controlled by few filters \cite{chen17} or parameters in the network \cite{dumoulin17}. It thus becomes possible to do a training for several styles jointly. At runtime, an image can then be transformed to any of the preselected styles by choosing the corresponding set of filters or parameters. Finally, recent works show that it is possible to train a deep neural network that takes as inputs any pair of style and content images and produces a stylized image, even for style images not viewed at training time \cite{li17,huang17,ghiasi17}. Furthermore, it was also showed that it is possible to control the level of stylisation or to mix different styles by doing a linear interpolation between the parameters of the deep network that control the corresponding styles. 

While the systems described so far provide impressive results when the style is a painting, the results are less impressive when the style is a natural image. Another line of research concerns the adaptation of the previous techniques to perform photorealistic style transfer. In~\cite{luan17}, the authors adapt the loss of~\cite{gatys16} with an additional term that favours photorealistic results. This additional term ensures a better preservation of the geometry of the content image. As in \cite{gatys16}, this new method is iterative and slow. Up until relatively recently, no method providing a fast solution of this photorealistic style transfer problem existed. A technical report proposing a fast method has however been made public \cite{li18} while writing this report. We discuss this alternative method in Section~\ref{sec:conclusion}.

\textbf{Contributions.} First, we propose a new network architecture to perform fast style transfer. The architecture is directly derived from the structure of the gradient descent algorithm used in \cite{gatys16} to solve the style transfer problem. We confirm using experiments that this network can provide fast artistic style transfer results. Second, we show that the network's architecture is flexible enough to be modified at runtime -- thus \textit{with no retraining} -- to incorporate modification of the initial training loss function. Third, we propose a slight modification of the loss function proposed by~\cite{luan17} to perform photorealistic style transfer and modify our trained network accordingly. Fourth, we show that we achieve results similar to those obtain by~\cite{luan17} with a better preservation of fine details. Finally, we show that our network can be used to perform fast videorealistic style transfer.

\section{Related Work}
\textbf{Visual style transfer.} Example-based image editing, whereby an input image is modified so as to acquire some of the characteristics of a reference picture, has a long history, \eg\ \cite{reinhard2001color,efros2001image,hertzmann2001image} for textures and colors. Recently, the specific problem of example-based stylization of a photograph, with the reference being typically a painting, has been revisited with great success by Gatys \etal\ \cite{gatys16}, using pre-trained deep convolutional features as the key ingredient. Through iterative optimization, a new image is created \textit{ex-nihilo} such that it retains the main structure (the ``content'') of the original photo while adopting the look and feel (colors, textures, paint strokes, the ``style'' in short) of the reference. This two-fold goal is encapsulated in a cost function over deep features of the three images. One major limitation of the approach is the cost of the optimization, with each iteration requiring forward-backward computations through the deep convnet that extracts the feature maps. Several authors, starting with \cite{johnson2016perceptual}\cite{ulyanov2016texture}, have showed that another feed-forward deep convnet can be designed and trained without supervision, for one or several given target styles, to perform the task much faster with similar quality: given a new content, this image is directly transformed into the stylized content without resorting to costly optimization. Our approach is also of this type, \ie, unsupervised learning of a ``neural solver'' for the original optimization problem, but with a very different take on the design of the architecture. Namely, we rely on the powerful idea of unrolling an iterative process to turn it into a trainable neural network. One key advantage of our approach is that important modifications of the original cost function can be mimicked in our architecture with no need for retraining. We demonstrate this in the case of photorealistic transfer \cite{luan17}. Fast methods in \cite{johnson2016perceptual,ulyanov2016texture,chen17,dumoulin17} might struggle with this extension of Gatys \etal's cost function, in particular due to the matting Laplacian that plays a crucial role, and would require retraining in any case. In contrast, a simple runtime restructuring of our architecture can accommodate these major changes. Note that alternative optimisation-based techniques to \cite{luan17}, such as \cite{he17}, exist for photorealistic color/style transfer. We use our flexible framework to rapidly solve \cite{luan17} but the principles detailled here could be used to solve other optimisation problems. Finally, we discuss in Section~\ref{sec:conclusion} the fast photorealistic style transfer method \cite{li18} which was made public while writing this report.

\textbf{Neural nets from algorithms unrolling.} A wide range of classic iterative solvers, \eg, to solve optimization problems or partial differential equations, amount to the repeated application of linear and component-wise non-linear maps. Unfolding such an algorithm over a fixed number of iterations allows one to see it as several layers of a neural network with shared, predefined weights. This can be taken as the starting point for an actual, trainable neural net. This powerful idea was introduced by Gregor and LeCun \cite{gregor2010learning} in the context of sparse coding: the popular iterative shrinkage thresholding algorithm (ISTA) \cite{beck2009fast} gives rise of a learnable network (LISTA) for fast approximate sparse coding. Each layer has independent trainable weights and the whole network is trained under the supervision of ISTA solutions. In the related context of linear inverse problems, including compressed sensing \cite{mousavi2017learning,metzler2017learned,borgerding2017amp,zhang2017ista} and image restoration \cite{wang2016proximal,liu2017proximal,chen2017trainable}, several works have followed this unrolling approach. In some cases, weight tying across layers is maintained as in the unrolled algorithm \cite{wang2016proximal}. In some other cases, the architecture departs notably from the unrolled algorithm, \eg, through using multiple convolutional filters per layer instead of only one as suggested by the unrolled iteration \cite{chen2017trainable}. 
Let us stress out that, by essence of such inverse problems, full supervision is possible: input-output samples of the transform to be inverted are indeed readily obtained, \eg, by sensing or artificially degrading natural images. By contrast, our approach is fully unsupervised, as other fast style transfer alternatives. In the absence of known outputs for corresponding input training samples, the training loss is defined as the original cost function. 

\begin{figure}[t]
\centering
\includegraphics[width=.70\linewidth]{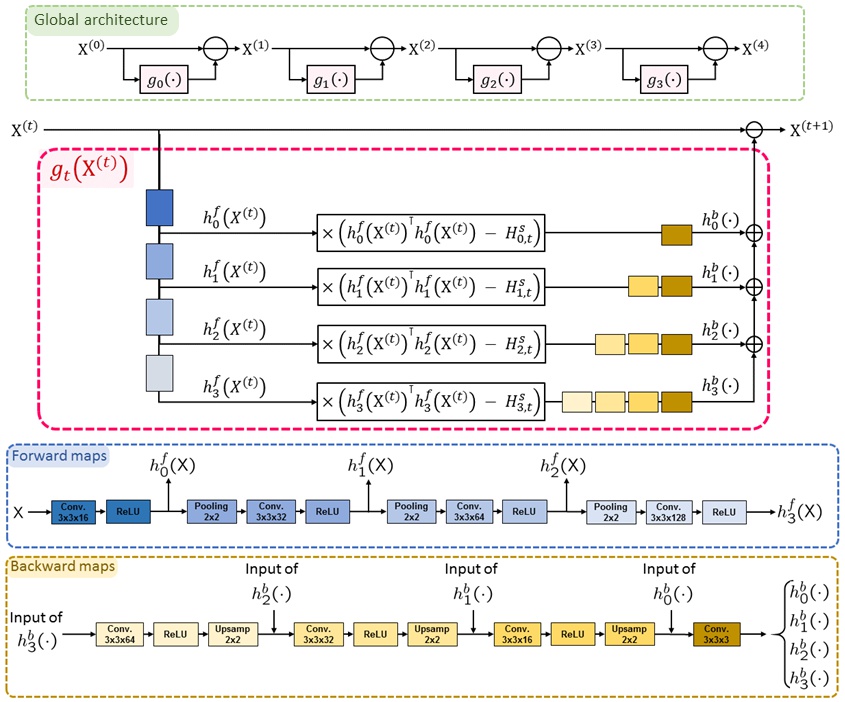}
\caption{\label{fig:combined_architectures}First (top): Global structure of the trained convnet for fast style transfer. It is a stack of four networks with identical structures, each one transforming an input image into a new one through one step of learned descent. Second: The structure of the descent maps $\{ g_t \}$. The coloured blocks correspond to sub-networks with same colors in the forward and backward maps $\{h^f_{\ell}\}$ and $\{h^b_{\ell}\}$. Third and fourth: Structure of the forward and backward maps for the computation of $\{ g_t \}$.}
\end{figure}
%

\textbf{Unsupervised neural solvers.} Training without supervision a (fast) feed-forward network to solve approximately a complex optimization problem is a challenging problem, with important applications. Using the cost function of interest as training loss seems straightforward, but the capacity of the designed architecture and the amount of actual guidance provided by the loss must be sufficient for such an unsupervised training to succeed. Besides works in fast style transfer, other recent works have demonstrated successful instances of this paradigm in other contexts. In \cite{tompson2016accelerating}, efficient fluid simulation is addressed. Within each time step of a classic solver of the incompressible Euler equation, the costly solving of the Poisson equation on the pressure field is replaced by a trained 3D convnet. In this case the loss amounts to the divergence square norm. In a very different context, \cite{tewari2017mofa} learn a deep neural solver for a complex inverse rendering problem: estimating the shape, expression and reflectance of a face in a single image, by minimization of the pixel-wise discrepancy between rendered and real faces. In both works, trained solvers provide good quality approximation of the solutions of the initial minimisation problem, but with massive acceleration compared to iterative solvers, and with no need for initialization in the latter work. In the present paper, we also propose to learn an unsupervised neural solver for a complex optimization problem, but following the unrolling principle explained above. As a consequence, the acceleration, already obtained by others on the same problem, is not our main contribution. More importantly, the unrolling approach gives the possibility to adapt at runtime our network in order to accommodate important changes to the original cost function.

\section{Network Architecture}
\subsection{Style Transfer by Gradient Descent}

Style transfer consists in transforming an image $\Sig_c \in \Rbb^{n \times 3}$ -- of $n$ pixels and $3$ color channels\footnote{Throughout, the two spatial dimensions of images and features maps are flattened into a single vector dimension for notational convenience.} -- to give it the ``style'' of another image $\Sig_s \in \Rbb^{n' \times 3}$ while preserving the content in $\Sig_c$. The style transfer method of Gatys \emph{et al.} \cite{gatys16} consists in solving a minimisation problem involving deep features obtained from a pre-trained convnet, typically the VGG-19 network \cite{simonyan14}, to obtained a  stylised image $\Sig \in \Rbb^{n \times 3}$. We denote by $\ma{F}_\ell, \ma{C}_\ell \in \Rbb^{{n_\ell} \times c_\ell}$ and $\ma{S}_{\ell} \in \Rbb^{{n'_\ell} \times c_\ell}$ the $\ell^\th$ feature maps -- in the VGG-19 network -- obtained from the images $\Sig, \Sig_c$ and $\Sig_s$, respectively. The minimised loss takes the form
\begin{align}
\label{eq:style_transfer_loss_global}
\set{L}(\Sig) = \lambda_c \, \set{L}_c(\Sig, \Sig_c) + \lambda_s \, \set{L}_s(\Sig, \Sig_s),
\end{align}
where the content loss $\set{L}_c(\Sig) = \abs{\mathcal{I}_c}^{-1} \; \sum_{\ell \in \mathcal{I}_c} \norm{\ma{F}_\ell - \ma{C}_\ell}_{\rm F}^2/(n_\ell \, c_\ell)$ ensures transfer of the content of $\Sig_c$ in the final image ($\norm{\cdot}_{\rm F}$ stands for matrix Frobenius norm), while the style loss
\begin{align}
\set{L}_s(\Sig) = \frac{1}{\abs{\mathcal{I}_s}} \; \sum_{\ell \in \mathcal{I}_s}  \frac{1}{c_\ell^2} \, \norm{ \frac{1}{n_\ell} \, \ma{F}_\ell^\adjoint \ma{F}_\ell - \frac{1}{n'_\ell} \, \ma{S}_\ell^\adjoint \ma{S}_\ell}_{\rm F}^2
\end{align}
ensures transfer of the style of $\Sig_s$ to the final image, and $\lambda_c, \lambda_s>0$. As in \cite{gatys16}, the layers used to encode the style and the content are $\set{I}_s = \{\text{conv}_{1\_1}$, $\text{conv}_{2\_1}$, $\text{conv}_{3\_1}$, $\text{conv}_{4\_1}$, $\text{conv}_{5\_1} \}$ and $\set{I}_c = \{\text{conv}_{4\_2} \}$, respectively. 

A stylised image $\Sig^*$ is usually obtained by starting from a random image $\Sig^{(0)}$ and by progressively updating it to minimise the loss $\set{L}$ by gradient descent:
\begin{align}
\label{eq:gradient_descent}
\Sig^{(t+1)} = \Sig^{(t)} - \mu \left[ \lambda_c \, \nabla\set{L}_c(\Sig^{(t)}) + \lambda_s \, \nabla\set{L}_s(\Sig^{(t)}) \right],
\end{align}
where $\mu > 0$ is the stepsize. While achieving impressive results, this method is nevertheless slow. To overcome this issue, researchers have designed deep convnets trained to solve \eqref{eq:style_transfer_loss_global} approximately at a much lower computational cost \cite{johnson2016perceptual,ulyanov2016texture,ghiasi17}, that is a neural map $f(\cdot)$ such that $\set{L}\big(f(\Sig^{(0)})\big) \approx \min_{\Sig} \set{L}(\Sig)$ provided that $\Sig_c$ is a natural image and that the network input is suitable, \eg, $\Sig^{(0)}=\Sig_c$ in \cite{johnson2016perceptual,ghiasi17} or a combination of $\Sig_c$ and additive white noise in \cite{ulyanov2016texture}.

\subsection{Learned Gradient Descent for Style Transfer}

We propose a new deep neural network to achieve fast style transfer. The architecture of this network is directly inspired by the structure of the gradient descent algorithm, following the idea of unrolling \cite{gregor2010learning}. This structure gives a better interpretability to the role of each filter as they correspond to specific parts of the original gradient descent algorithm. This property permits us to augment, at runtime, the network with new desirable features, for instance to perform photorealistic style transfer.

\noindent\textbf{Global architecture.} The proposed convnet follows the update rule of the gradient descent algorithm \eqref{eq:gradient_descent} where the actual gradient is replaced by a learned descent direction:\footnote{There is no mathematical guarantee that $g_{t} \big( \Sig^{(t)} \big)$ is indeed a descent direction.}
\begin{align}
\label{eq:network_global_structure}
\Sig^{(t+1)} = \Sig^{(t)} - g_{t} \left( \Sig^{(t)} \right), 
\quad
t = 0, 1, 2, 3.
\end{align}
The global architecture of the network is illustrated in Fig.~\ref{fig:combined_architectures}. Note that the proposed network can be viewed as a residual network \cite{he_resnet16}. We restrict the computation to $4$ iterations to maintain the computational cost low and constant. By comparing \eqref{eq:gradient_descent} to \eqref{eq:network_global_structure}, we see that $g_{t} \left( \Sig^{(t)} \right)$ plays the role of the gradient. We explain next how $g_{t} \left( \Sig^{(t)} \right)$ is computed.

\noindent\textbf{Computing the descent direction.} The gradient in \eqref{eq:gradient_descent} is made of a content term and a style term. To simplify the network and reduce computational costs, we design $g_{t}$ by mimicking the computational architecture of $\nabla\set{L}_s$ only. However, as in \cite{johnson2016perceptual}, we initialise $\Sig^{(0)}$ with the RGB image\footnote{At training time, a zero-mean uniform noise of amplitude drawn uniformly in $[0, 0.1]$ is added to $\Sig^{(0)}$.} $\Sig_c$ -- after scaling the pixel values in $[0, 1]$ -- instead of the classic random initialization of the gradient descent, and we use the complete cost $\set{L}$ as training loss. This allows us to overcome the lack of $\nabla\set{L}_c$ in the network and keep the content of $\Sig_c$ in the final image $\Sig^{(4)}$. 

Let us now concentrate on $\nabla\set{L}_s$. We recall that
\begin{align}
\label{eq:partial_derivative_style}
\frac{\partial\set{L}_s}{\partial \ma{F}_\ell} \propto \ma{F}_\ell \cdot \left[ \frac{1}{n_\ell} \, \ma{F}_\ell^\adjoint \ma{F}_\ell - \frac{1}{n'_\ell} \, \ma{S}_\ell^\adjoint \ma{S}_\ell \right].
\end{align}
The role of the difference of Gram matrices in \eqref{eq:partial_derivative_style} is to correct the feature maps $\ma{F}_\ell$ so that, after backpropagation, the image $\Sig^{(t)}$ gets closer to the style of $\Sig_s$. The filter resulting from the difference of Gram matrices is called \emph{style correction filter} hereafter. The gradient of $\set{L}_s$ is obtained by
\begin{enumerate}[(i)]
\item\label{it:forward} computing the feature maps $\ma{F}_\ell$ (forward maps); 
\item\label{it:partial_derivative} computing the partial derivatives \eqref{eq:partial_derivative_style}; 
\item\label{it:backward} back-propagating these partial derivatives to the input (backward maps). 
\end{enumerate}
We construct $g_{t}$ by mimicking these $3$ steps with a simpler convnet than VGG-19. 

[\textit{Transforming Steps (\ref{it:forward}) and (\ref{it:backward}})] The structure of the simplified forward and backward maps are presented in Fig.~\ref{fig:combined_architectures}. The new feature maps playing the role of $\{\ma{F}_\ell \}$ are $ \{ h_{\ell}^f(\Sig) \}$. As in the original style loss, the spatial dimension between two consecutive layers is divided by $2$ in both directions and that the number of feature channels is multiplied by $2$. Let us highlight that we use only $4$ layers instead of $5$ as originally in $\set{L}_s$. The reason is only computational. We noticed that this was enough to yield satisfactory results. The backward maps $ \{ h_{\ell}^b(\cdot) \}$ are obtained from the forward maps by symmetry. Let us highlight that the last layer of the backward maps does not involve any ReLU to allow negative descent direction. The pooling layers use average pooling and the upsampling layers use bilinear upsampling. All convolutions are computed using reflection padding. All the filters in these maps are learned and are \emph{identical} for all styles.

[\textit{Transforming Step (\ref{it:partial_derivative}})] The computation of the partial derivatives \eqref{eq:partial_derivative_style} is directly transformed to  
\begin{align}
\label{eq:estimated_pd}
 h_{\ell}^f (\Sig^{(t)}) \cdot \left[ \frac{1}{n_\ell} \, h_{\ell}^f (\Sig^{(t)})^\adjoint \, h_{\ell}^f (\Sig^{(t)})
\; - \; \ma{H}_{\ell, t}^s, \right]
\end{align}
where the learned matrix $\ma{H}_{\ell, t}^s$ plays the role of the Gram matrix $\ma{S}_\ell^\adjoint \ma{S}_\ell$ that encodes the style. Note however that the matrix $\ma{H}_{\ell, t}^s$ is different at each iteration $t$, unlike in \eqref{eq:partial_derivative_style}. The matrices $\{ \ma{H}_{\ell, t}^s \}$ are specific to each style. For simplicity, we do not force $\ma{H}_{\ell, t}^s$ to be symmetric and positive semi-definite during training. 

The complete structure of $g_{t}$ is presented in Fig.~\ref{fig:combined_architectures}. Let us highlight that the maps $\{ h^f_{\ell} \}$ and $\{ h^b_{\ell} \}$ are independent of $t$; they are shared across the four iteration networks. Note that the matrix multiplication with \eqref{eq:estimated_pd} can be implemented as a convolution with a filter of spatial size $1 \times 1$, so that the whole network is indeed convolutional. We also notice that this filter is instance dependent: its weights depend on $h_{\ell}^f (\Sig^{(t)})$. The forward and backward maps $\{h^f_{\ell}\}$ and $\{h^b_{\ell}\}$ contain $194\,755$ weights to learn. At each iteration $t$, the style matrices contain $21\,760$ trainable weights. The total number of parameters is thus $194\,755 + 21\,760 \cdot 4 = 281\,795$.

\noindent\textbf{Training.} Following \cite{johnson2016perceptual}, we augment $\set{L}$ defined in \eqref{eq:style_transfer_loss_global} with a Total Variation regularisation term to promote piecewise smoothness of the transformed images:
%
\begin{align}
\label{eq:training_cost}
\lambda_c \, \set{L}_c(\Sig^{(4)}) + \lambda_s \, \set{L}_s(\Sig^{(4)}) +  \lambda_{\rm TV} \,{\rm TV}(\Sig^{(4)}),
\end{align}
where the values in the RGB image $\Sig^{(4)}$ are clipped between $0$ and $1$. All filters in $ \{ h_{\ell}^f \}$, $\{ h_{\ell}^b \}$ are initialised using Xavier initialisation \cite{glorot10} and the matrices $\ma{H}_{\ell, t}^s$ are initialised at zero. The filters are trained using Adam \cite{kingma14} with a constant stepsize of $10^{-5}$ and a batchsize of $1$. The network is trained for $17$ epochs over the 2014 MS-COCO training dataset~\cite{lin14}.\footnote{\url{http://cocodataset.org/}} The training images are cropped to the largest possible square, resized to $320 \times 320$, and used as content images for training. We use $69$ style images in total, including $18$ images of paintings. The remaining $51$ images are photographs used by the authors of~\cite{luan17} and for which the masks to segment the images into different semantic regions are available.\footnote{\url{https://github.com/luanfujun/deep-photo-styletransfer}} Each semantic region of a photograph defines a style. We thus consider $153$ styles in total, coming from either paintings or photographs. The style term $\set{L}_s$ is therefore modified with semantic masks that apply on the style images during training. Let $\ma{M}^s \in \Rbb^{n' \times n'}$ be the binary diagonal matrix that encodes a semantic mask on the style image $\Sig_s$, the style term becomes 
\begin{align}
\set{L}_s(\Sig) = \frac{1}{\abs{\mathcal{I}_s}} \; \sum_{\ell \in \mathcal{I}_s}  \frac{1}{c_\ell^2} \, \norm{ \frac{1}{n_\ell} \, \ma{F}_\ell^\adjoint \ma{F}_\ell - \frac{1}{{\rm Tr}(\ma{M}^s_\ell)} \, (\ma{M}^s_\ell \ma{S}_\ell)^\adjoint (\ma{M}^s_\ell \ma{S}_\ell)}_{\rm F}^2
\end{align}
where $\ma{M}^s_\ell$ is the semantic mask propagated at the $\ell^\th$ layer of the VGG-19. Let us remark that in absence of semantic mask, then $\ma{M}^s_\ell$ is the identity, ${\rm Tr}(\ma{M}^s_\ell) = n'_\ell$, and we recover the original style loss. Note that the content images are not segmented during training. We use $\lambda_c = 0.025$, $\lambda_{\rm TV} = 0.5$, and 
\begin{align}
\lambda_{s} = \left[ \frac{1}{\abs{\mathcal{I}_s}} \; \sum_{\ell \in \mathcal{I}_s}  \frac{1}{c_\ell^2} \, \norm{\frac{1}{{\rm Tr}(\ma{M}^s_\ell)} \, (\ma{M}^s_\ell \ma{S}_\ell)^\adjoint (\ma{M}^s_\ell \ma{S}_\ell)}_{\rm F}^2 \right]^{-1}.
\end{align}
We notice that this choice of $\lambda_s$ allows similar degree of stylization across learned styles. The evolution of the training loss computed on $20$ images of the 2014 MS-COCO validation set is presented in Fig.~\ref{fig:fb_network_graph}. We remark that the loss starts to stabilise after around $14$ epochs.

\section{Runtime Restructuring}
In this section, we use properties of our trained network to obtain photorealistic results via modifications made at runtime. We recall what changes need to me made to the original loss function for style transfer to obtain photorealistic results. These changes directly impact the gradient descent algorithm used to solve this problem. As our network has the same structure as this algorithm, it is straightforward to make the same modifications in the network to obtain the desired results with no retraining.

\vspace{-2mm}
\subsection{Laplacian Regularisation for Photorealism}

\begin{figure}[t]
\centering
\includegraphics[width=\linewidth]{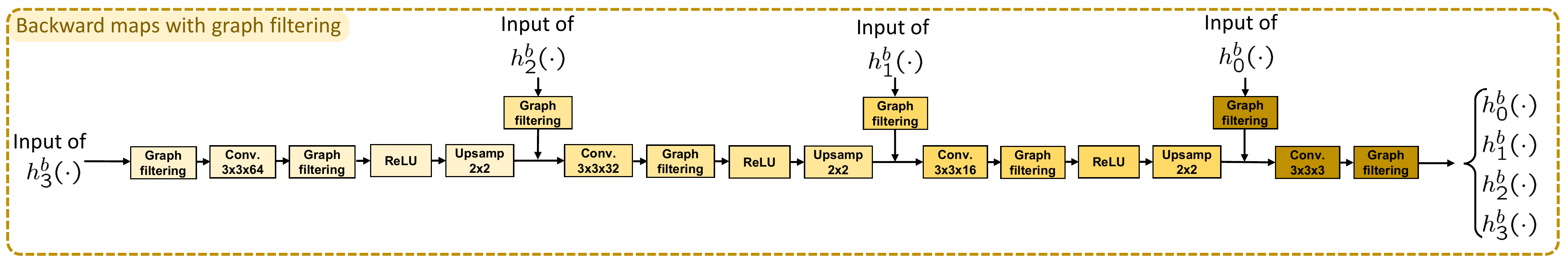}
\includegraphics[clip, trim=18mm 93mm 18mm 109mm, width=.32\linewidth]{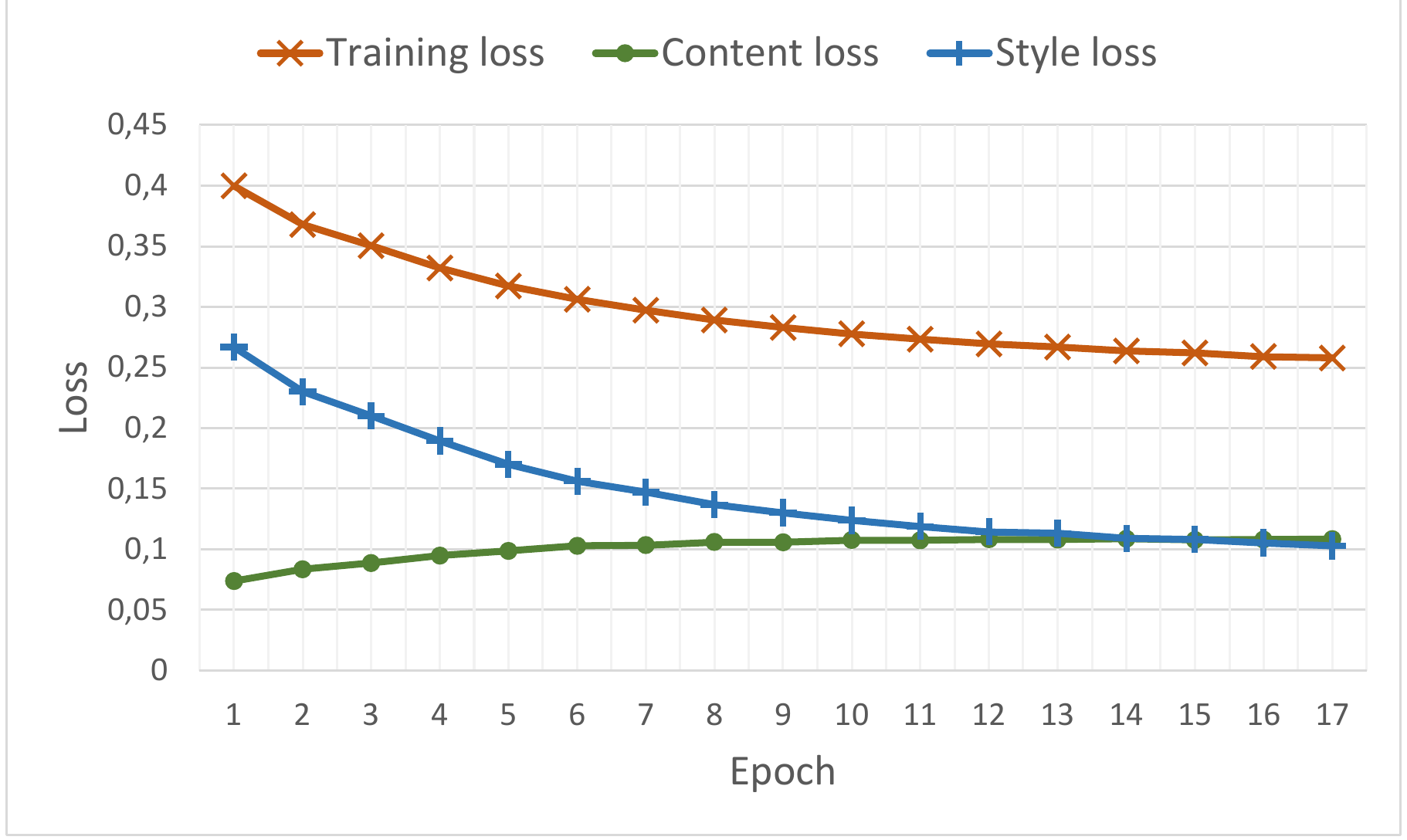}
\includegraphics[clip, trim=18mm 25mm 17mm 25mm, width=.32\linewidth]{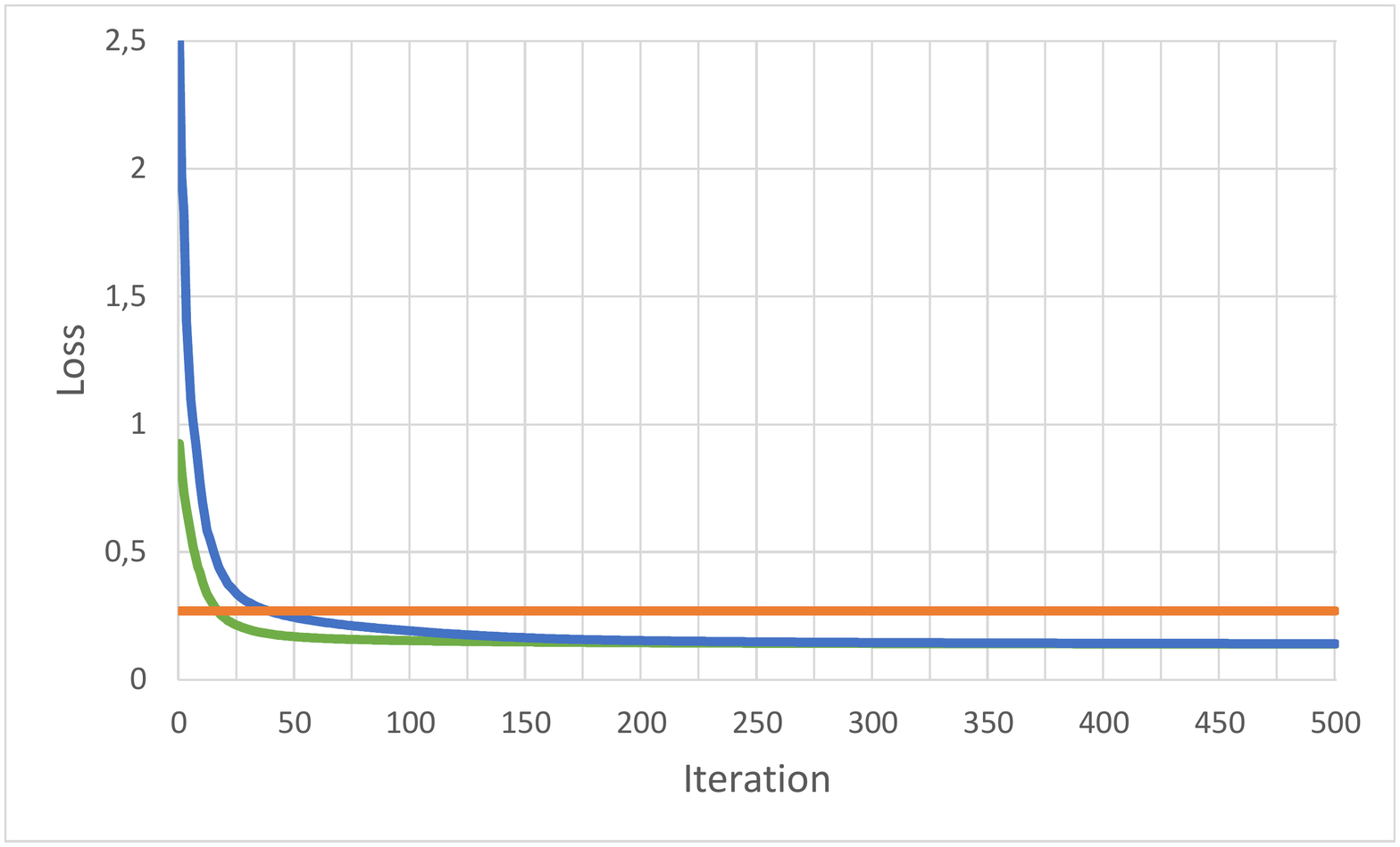}
\includegraphics[clip, trim=18mm 25mm 17mm 25mm, width=.32\linewidth]{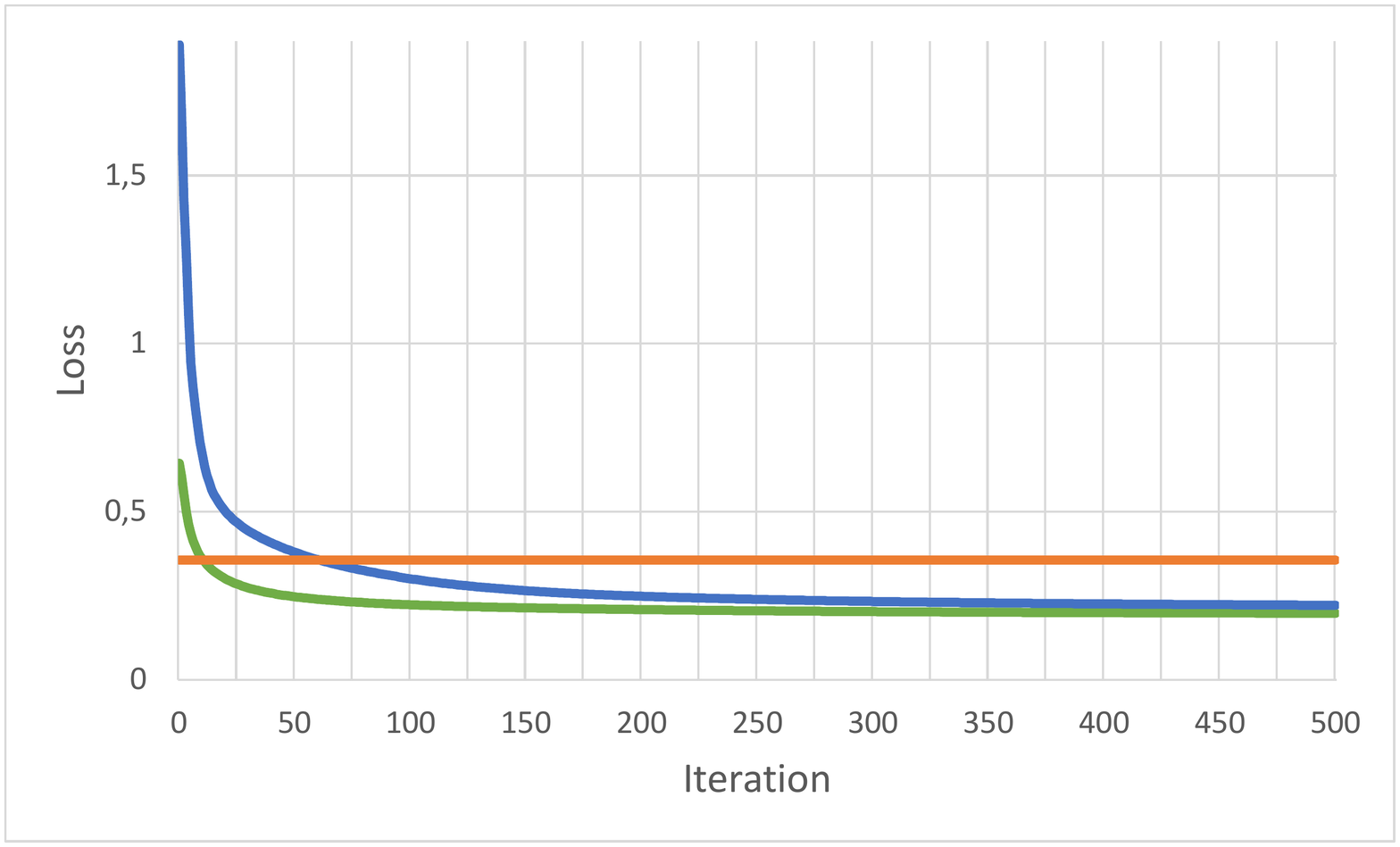}
\\
\caption{\label{fig:fb_network_graph}Top: Structure of the backward maps modified at runtime with graph filtering. Bottom left: Evolution of the training loss (orange), and of its summands related to $\set{L}_c$ (green) and $\set{L}_s$ (blue), on a validation set of $20$ images as a function of the number of epochs. Bottom, middle and right: Evolution of \eqref{eq:training_cost} vs. the number of iterations of L-BFGS when initialised with noise (blue) or $\Sig_c$ (green). Value of \eqref{eq:training_cost} attained with the solution of our trained network (orange). Curves are obtained using an image of \emph{The scream} (middle) or of the \emph{Portrait of Dora Maar} (right) as styles.}
\end{figure}

Minimizing \eqref{eq:style_transfer_loss_global} yields very good results when transferring the style of a painting to a photograph \cite{gatys16}. Unfortunately, the results are less impressive when transferring the style of a photograph to another photograph as the result is often not photorealistic. The authors of \cite{luan17} propose to solve this issue by favouring transformations from the content image to the final image that are locally affine. In practice, this is done by adding a penalty term to $\set{L}$, which becomes
\begin{align}
\label{eq:style_transfer_loss_global_photoreal}
\set{L}(\Sig) + \lambda_{\rm L} \, {\rm Tr} (\Sig^\adjoint \Lap \Sig),
\end{align}
where $\Lap \in \Rbb^{n \times n}$ is the Matting Laplacian \cite{levin08} of the content image $\Sig_c$. Even though our network was trained without this Laplacian term, we can enforce an equivalent constraint at runtime to obtain the desired results.

In graph signal processing terms, finding an image $\Sig$ for which ${\rm Tr} (\Sig^\adjoint \Lap \Sig)$ is small means finding an image that is smooth on the graph $\Graph$ encoded by the Laplacian matrix $\Lap$. Let us recall the concept of smoothness on graphs.

\noindent\textbf{Smooth graph signals.} The graph Fourier transform of any graph signal $\vec{x} \in \Rbb^{n}$ is defined as $\hat{\vec{x}} = \Fou^\adjoint \vec{x}$, where $\Fou = (\fou_1, \ldots, \fou_{n}) \in \Rbb^{n \times n}$ is the matrix of eigenvectors of $\Lap$: $\Lap = \Fou \Eig \Fou^\adjoint$, where $\Eig = \diag(\eig_1, \ldots, \eig_{n}) \in \Rbb^{n \times n}$ and $\eig_1 \leq \ldots \leq \eig_{n}$ \cite{shuman_SPMAG2013}. A smooth graph signal is a signal whose energy is concentrated on the first few Fourier coefficients (Fourier coefficients at the lowest graph frequency). A $k$-bandlimited graph signal is a signal whose energy is entirely concentrated on the first $k$ Fourier coefficients, and is thus a special type of smooth signal when $k \ll n$. The set of $k$-bandlimited graph signals is denoted ${\rm span}(\Fou_k)$ where $\Fou_k = (\fou_1, \ldots, \fou_k) \in \Rbb^{n \times k}$ is the restriction of $\Fou$ to its first $k$ columns. 

Another fundamental tool in graph signal processing is filtering. One can define spectral graph filtering as follows
$\vec{x} * u = \Fou \, \diag(u(\eig_1), \ldots, u(\eig_{n})) \, \Fou^\adjoint \, \vec{x}$, where $\vec{x} * u$ is the signal $\vec{x}$ filtered by $u : \Rbb_+ \rightarrow \Rbb$. Projecting a signal $\vec{x}$ onto the set of $k$-bandlimited graph signal is done by filtering with
\begin{align}
i_{\lambda^*} (\lambda)
= 
\left\{
\begin{array}{ll}
1 & \text{if } \lambda \leq  \lambda^*, \\
0 & \text{oth,}
\end{array}
\right.\quad
\text{with $\lambda^* \in [\lambda_{k}, \lambda_{k+1})$.}
\end{align}
%

\noindent\textbf{Constrained Version of \eqref{eq:style_transfer_loss_global_photoreal}.} 
Minimising ${\rm Tr} (\Sig^\adjoint \Lap \Sig)$ as in \eqref{eq:style_transfer_loss_global_photoreal} means finding an image that is smooth on the graph $\Graph$ encoded by the Laplacian matrix $\Lap$. Instead of adding the penalisation term ${\rm Tr} (\Sig^\adjoint \Lap \Sig)$ to $\set{L}$, one can also search for an image $\Sig$ smooth on $\Graph$ by solving the constrained problem
\begin{align}
\label{eq:style_transfer_loss_global_photoreal_constrained}
\min_{\Sig} \set{L}(\Sig) 
\quad
\text{ subject to }
\quad
\Sig \in {\rm span}(\Fou_k).
\end{align}
This problem can be solved by projected gradient descent whose iterations satisfy:
\begin{align}
\Sig^{(t+1)} = \left[\Sig^{(t)} - \mu \, \nabla \set{L} \left( \Sig^{(t)} \right) \right] * i_{\lambda^*},
\end{align}
where the filtering with $i_{\lambda^*}$ projects the iterates onto the space of $k$-bandlimited signals on $\Graph$. The only difference with \eqref{eq:gradient_descent} is thus the extra projection after each gradient update. It is straightforward to add this constraint in our deep CNN by changing \eqref{eq:network_global_structure} to
\begin{align}
\Sig^{(t+1)} = \Sig^{(t)} - g_t \left( \Sig^{(t)} \right) * i_{\lambda^*}.
\end{align}
As we initialise $\Sig^{(0)}$ with $\Sig_c$, and because we can safely assume that $\Sig_c \approx \Sig_c * i_{\lambda^*}$ (otherwise the smoothness constraint would not be a good photorealistic prior), it is enough to filter $g_t \left( \Sig^{(t)} \right)$ at every iteration to ensure that all $\Sig^{(t)}$ are $k$-bandlimited.\footnote{Remark that the space of $k$-bandlimited signals is linear.} Adding this graph filtering step at runtime is sufficient to transform a network trained to solve \eqref{eq:style_transfer_loss_global_photoreal} to a new network solving \eqref{eq:style_transfer_loss_global_photoreal_constrained}.

\noindent\textbf{Fast graph filtering.} Filtering with $i_{\lambda^*}$ requires the computation of the matrix $\Fou$. To avoid this intractable computation, we use a polynomial approximation $\tilde{i}_{\lambda^*}$ of $i_{\lambda^*}$ to perform filtering. Polynomial filters are computationally efficient as they only require matrix vector multiplication with the sparse Laplacian matrix $\Lap$. We let the reader refer to, \eg, \cite{hammond11} for more explanations on this subject. In practice, we use Jackson-Chebychev polynomials \cite{napoli13} of order $5$ computed to approximate $i_{\lambda^*}$ with $\lambda^* = 0.2 \, \lambda_{n}$.\footnote{Fixing $k$ is also possible but requires additional computations to estimate $\lambda^*$.} Note that polynomial graph filters are also used to build efficient graph CNNs \cite{defferrard16,kipf16}.

We noticed that to achieve more pleasing results visually, it is better to filter $g_t \left( \Sig^{(t)} \right)$ at different resolutions. It permits to preserve better large scale and low scale structures. We take advantage of the multiscale structure of the trained network to compute this multiscale filtering efficiently. Each feature map after computing \eqref{eq:estimated_pd} and in the four levels of the backward pass is graph filtered individually (channelwise). This graph filtering is placed after each convolutional layer and before the ReLU, if present, as presented in Fig.~\ref{fig:fb_network_graph}. The four used Laplacian matrices are computed on the initial content image $\Sig_c$ downsampled at the corresponding resolutions of the feature maps in $\{ h^{b}_{\ell} \}$. Note that, for fair comparison, we post-process $\Sig^{(4)}$ using a guided filter as done in \cite{luan17,li18}. The effect of this post-processing is studied in the Appendix. It permits to remove artefacts that sometimes appear near edges, to sharpen some edges, but also to correct compression artefacts when present.

\begin{figure}[t]
\centering
\begin{minipage}{0.80\linewidth}
\begin{minipage}{0.24\linewidth}\centering $\Sig_c$ \end{minipage}
\begin{minipage}{0.24\linewidth}\centering \scriptsize \cite{gatys16} - Noise as init.\end{minipage}
\begin{minipage}{0.24\linewidth}\centering \scriptsize \cite{gatys16} - $\Sig_c$ as init. \end{minipage}
\begin{minipage}{0.24\linewidth}\centering \scriptsize Ours \end{minipage}
\\
\begin{minipage}{0.24\linewidth}
\includegraphics[width=\linewidth]{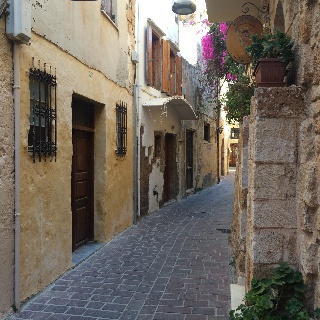}
\end{minipage}
\begin{minipage}{0.24\linewidth}
\includegraphics[width=\linewidth]{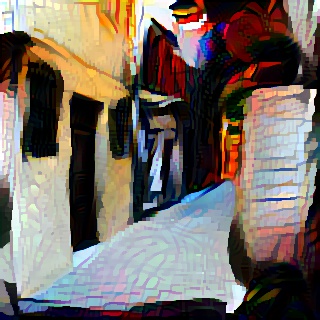}
\end{minipage}
\begin{minipage}{0.24\linewidth}
\includegraphics[width=\linewidth]{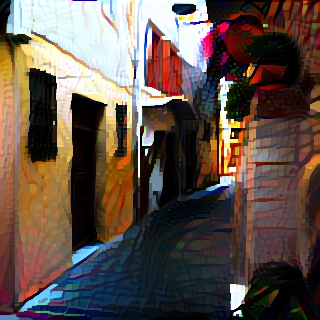}
\end{minipage}
\begin{minipage}{0.24\linewidth}
\includegraphics[width=\linewidth]{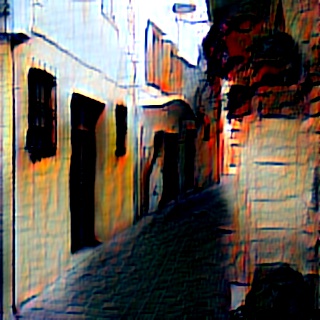}
\end{minipage}
\\
\begin{minipage}{0.24\linewidth}
\includegraphics[width=\linewidth]{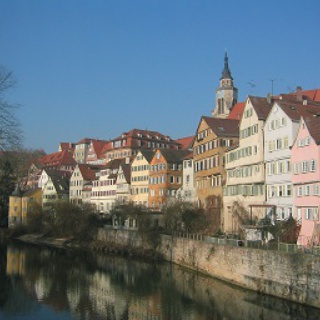}
\end{minipage}
\begin{minipage}{0.24\linewidth}
\includegraphics[width=\linewidth]{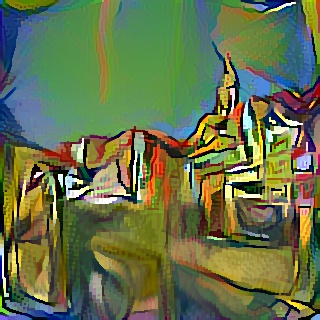}
\end{minipage}
\begin{minipage}{0.24\linewidth}
\includegraphics[width=\linewidth]{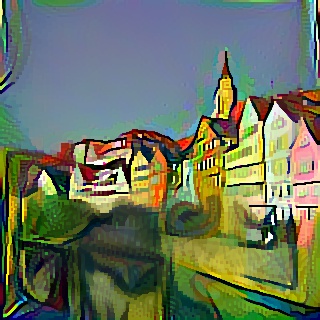}
\end{minipage}
\begin{minipage}{0.24\linewidth}
\includegraphics[width=\linewidth]{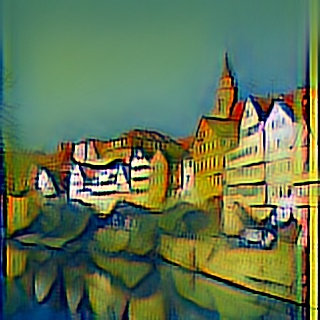}
\end{minipage}
\\
\begin{minipage}{0.24\linewidth}
\includegraphics[width=\linewidth]{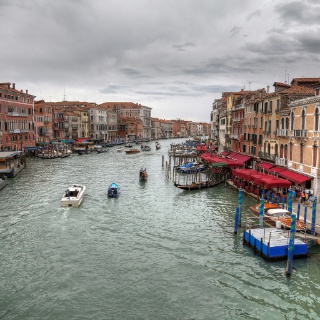}
\end{minipage}
\begin{minipage}{0.24\linewidth}
\includegraphics[width=\linewidth]{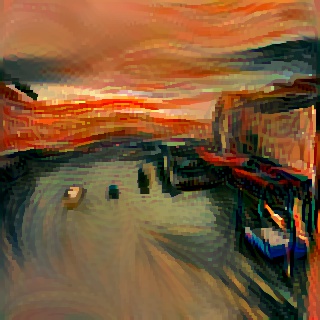}
\end{minipage}
\begin{minipage}{0.24\linewidth}
\includegraphics[width=\linewidth]{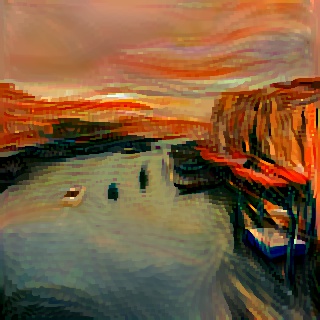}
\end{minipage}
\begin{minipage}{0.24\linewidth}
\includegraphics[width=\linewidth]{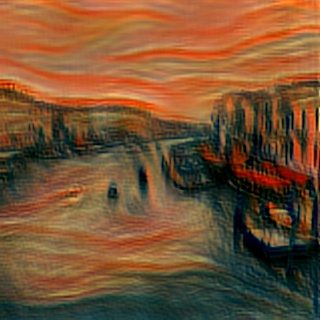}
\end{minipage}
\end{minipage}
\hfill
\vline
\hfill
\begin{minipage}{0.19\linewidth}
\begin{minipage}{\linewidth} 
\centering
\scriptsize
\textbf{Style 1},
\emph{Portrait of Dora Maar},
P.~Picasso, 1937.
\end{minipage}
\\
\begin{minipage}{\linewidth} 
\centering
\includegraphics[height=22mm]{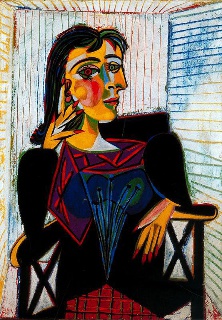}
\end{minipage}
\vspace{1mm}
\\
\begin{minipage}{\linewidth} 
\centering
\scriptsize
\textbf{Style 2},
\emph{A Muse},
P.~Picasso, 1935.
\end{minipage}
\\
\begin{minipage}{\linewidth} 
\centering
\includegraphics[width=0.94\linewidth]{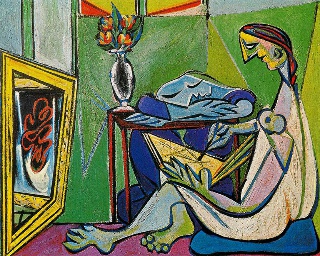}
\end{minipage}
\vspace{1mm}
\\
\begin{minipage}{\linewidth} 
\centering
\scriptsize
\textbf{Style 3},
\emph{The scream},
E.~Munch, 1893.
\end{minipage}
\\
\begin{minipage}{\linewidth} 
\centering
\includegraphics[height=22mm]{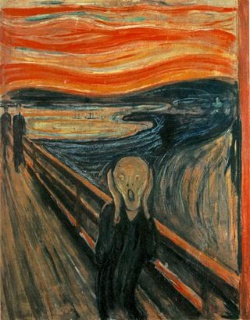}
\end{minipage}
\end{minipage}
\caption{\label{fig:artistic_style_transfer}Artistic style transfer results obtained with~\cite{gatys16} using two different initialisations (second and third columns) and our network (fourth column), for $3$ different content images (first column) and corresponding styles (last column).}
\end{figure}
%

\subsection{Masking at Runtime}

\begin{figure}[t]
\centering
\begin{minipage}{0.79\linewidth}
\begin{minipage}{0.24\linewidth}\centering \scriptsize $\Sig_c$ \end{minipage}
\begin{minipage}{0.24\linewidth}\centering \scriptsize \cite{luan17} \end{minipage}
\begin{minipage}{0.24\linewidth}\centering \scriptsize Ours - No filt.\end{minipage}
\begin{minipage}{0.24\linewidth}\centering \scriptsize Ours - Graph filt. \end{minipage}
\\
\begin{minipage}{0.24\linewidth}
\includegraphics[width=\linewidth]{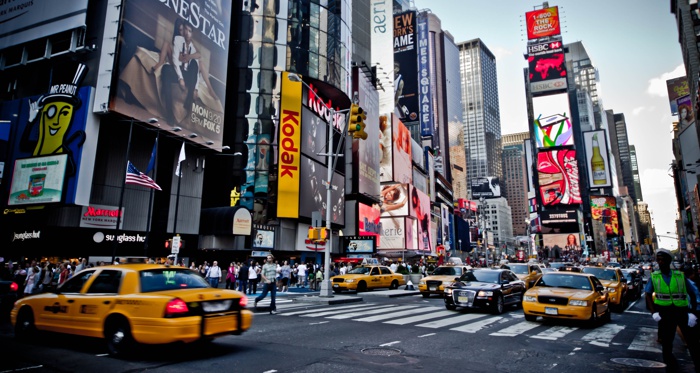}\\
\includegraphics[width=0.31\linewidth]{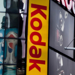}
\includegraphics[width=0.31\linewidth]{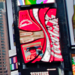}
\includegraphics[width=0.31\linewidth]{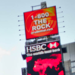}
\end{minipage}
\begin{minipage}{0.24\linewidth}
\includegraphics[width=\linewidth]{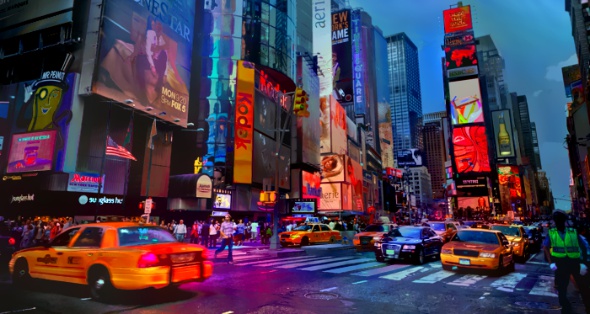}\\
\includegraphics[width=0.31\linewidth]{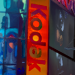}
\includegraphics[width=0.31\linewidth]{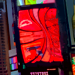}
\includegraphics[width=0.31\linewidth]{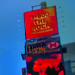}
\end{minipage}
\begin{minipage}{0.24\linewidth}
\includegraphics[width=\linewidth]{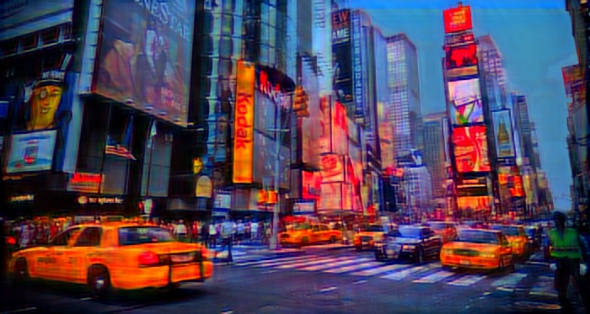}\\
\includegraphics[width=0.31\linewidth]{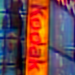}
\includegraphics[width=0.31\linewidth]{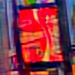}
\includegraphics[width=0.31\linewidth]{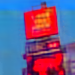}
\end{minipage}
\begin{minipage}{0.24\linewidth}
\includegraphics[width=\linewidth]{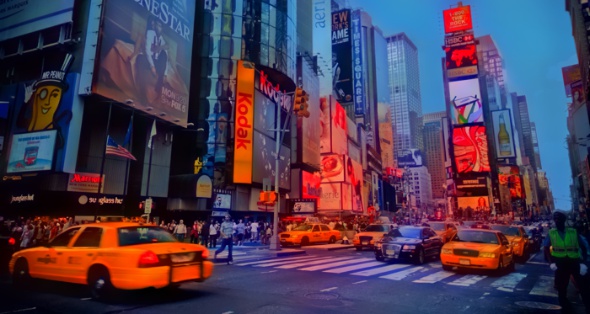}\\
\includegraphics[width=0.31\linewidth]{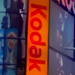}
\includegraphics[width=0.31\linewidth]{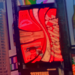}
\includegraphics[width=0.31\linewidth]{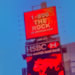}
\end{minipage}
\\
\begin{minipage}{0.24\linewidth}
\includegraphics[width=\linewidth]{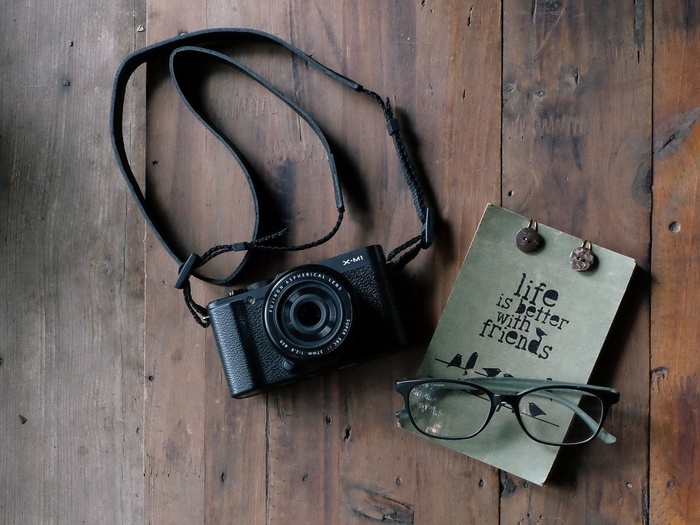}\\
\includegraphics[width=0.31\linewidth]{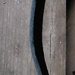}
\includegraphics[width=0.31\linewidth]{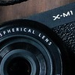}
\includegraphics[width=0.31\linewidth]{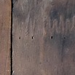}
\end{minipage}
\begin{minipage}{0.24\linewidth}
\includegraphics[width=\linewidth]{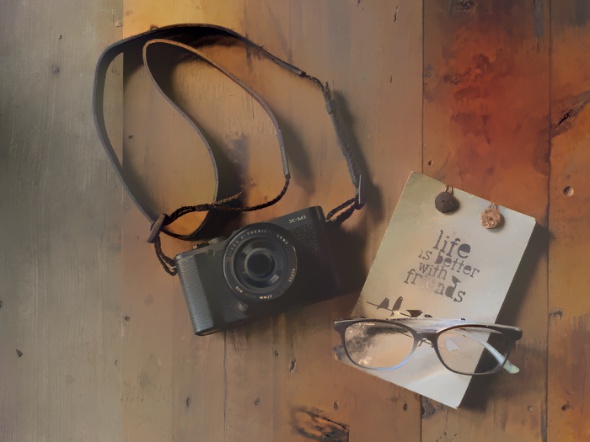}\\
\includegraphics[width=0.31\linewidth]{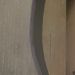}
\includegraphics[width=0.31\linewidth]{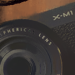}
\includegraphics[width=0.31\linewidth]{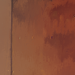}
\end{minipage}
\begin{minipage}{0.24\linewidth}
\includegraphics[width=\linewidth]{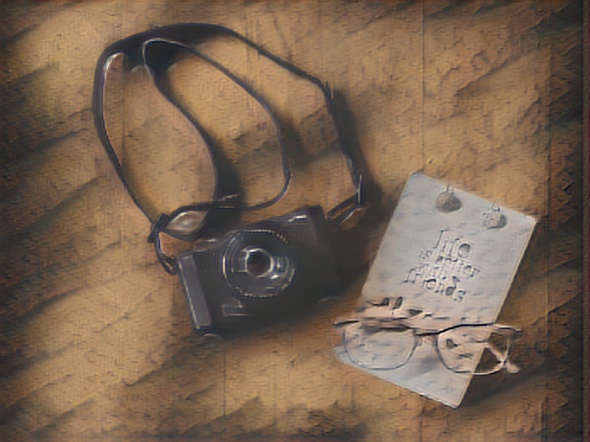}\\
\includegraphics[width=0.31\linewidth]{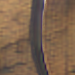}
\includegraphics[width=0.31\linewidth]{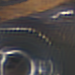}
\includegraphics[width=0.31\linewidth]{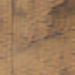}
\end{minipage}
\begin{minipage}{0.24\linewidth}
\includegraphics[width=\linewidth]{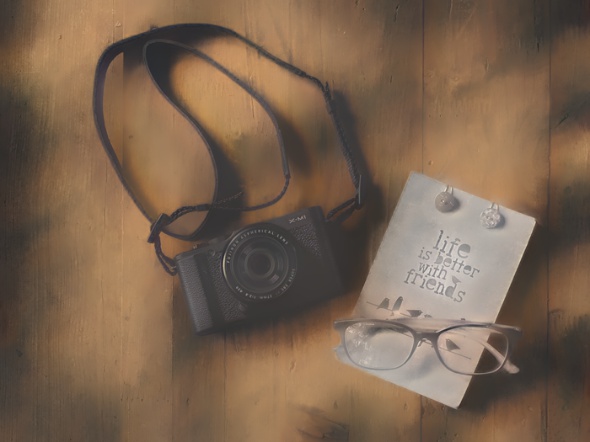}\\
\includegraphics[width=0.31\linewidth]{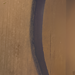}
\includegraphics[width=0.31\linewidth]{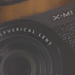}
\includegraphics[width=0.31\linewidth]{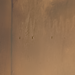}
\end{minipage}
\\
\begin{minipage}{0.24\linewidth}
\includegraphics[width=\linewidth]{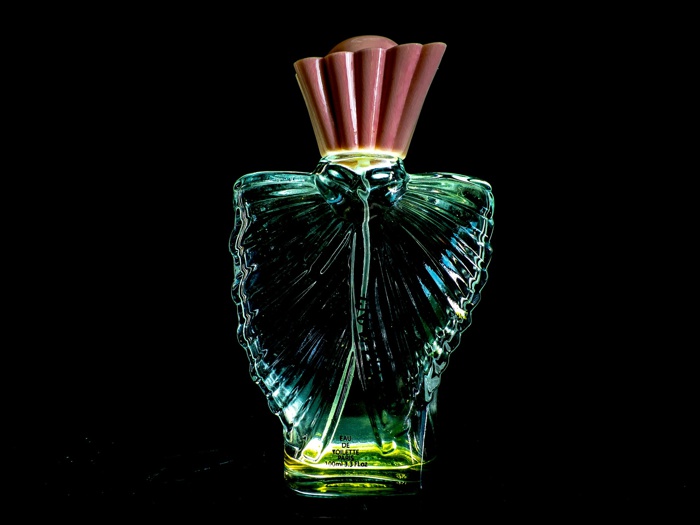}\\
\includegraphics[width=0.31\linewidth]{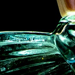}
\includegraphics[width=0.31\linewidth]{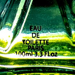}
\includegraphics[width=0.31\linewidth]{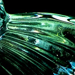}
\end{minipage}
\begin{minipage}{0.24\linewidth}
\includegraphics[width=\linewidth]{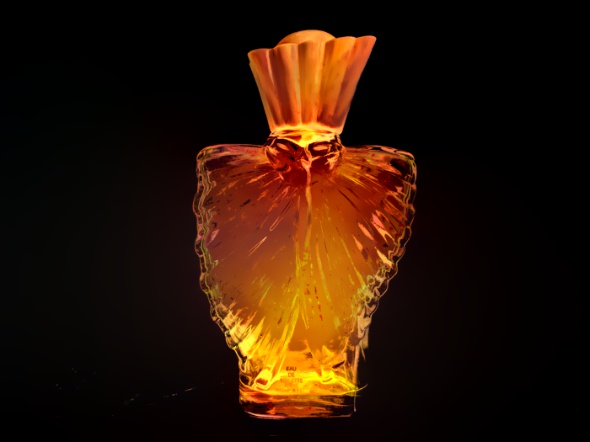}\\
\includegraphics[width=0.31\linewidth]{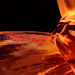}
\includegraphics[width=0.31\linewidth]{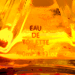}
\includegraphics[width=0.31\linewidth]{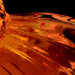}
\end{minipage}
\begin{minipage}{0.24\linewidth}
\includegraphics[width=\linewidth]{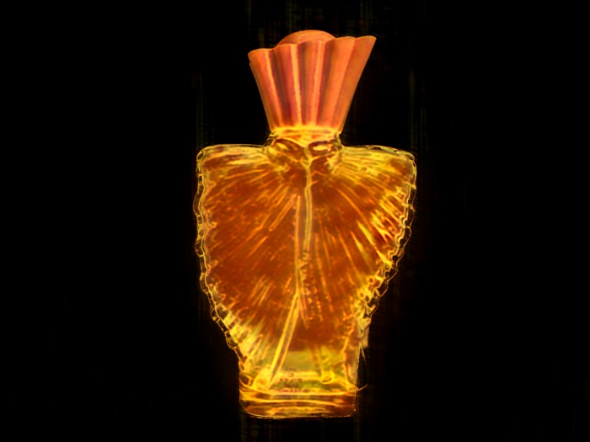}\\
\includegraphics[width=0.31\linewidth]{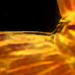}
\includegraphics[width=0.31\linewidth]{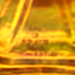}
\includegraphics[width=0.31\linewidth]{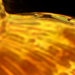}
\end{minipage}
\begin{minipage}{0.24\linewidth}
\includegraphics[width=\linewidth]{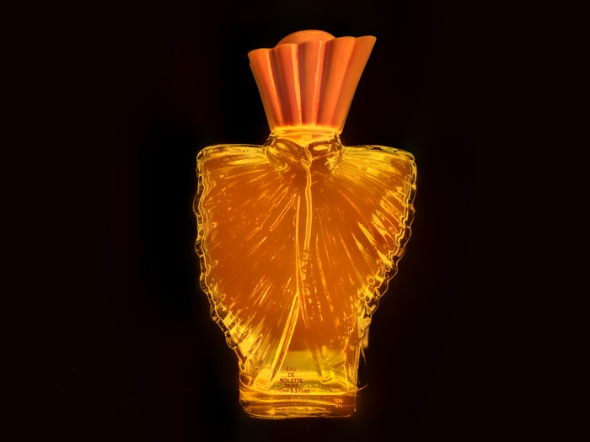}\\
\includegraphics[width=0.31\linewidth]{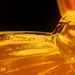}
\includegraphics[width=0.31\linewidth]{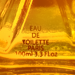}
\includegraphics[width=0.31\linewidth]{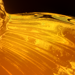}
\end{minipage}
\end{minipage}
\hfill
\vline
\hfill
\begin{minipage}{0.19\linewidth}
\begin{minipage}{\linewidth} 
\centering
\scriptsize
\textbf{Style 1}
\end{minipage}
\\
\begin{minipage}{\linewidth} 
\centering
\includegraphics[width=\linewidth]{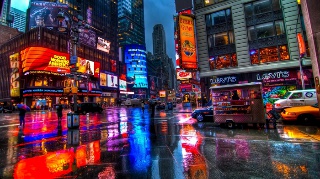}
\end{minipage}
\\
\vspace{3mm}
\\
\begin{minipage}{\linewidth} 
\centering
\scriptsize
\textbf{Style 2}
\end{minipage}
\\
\begin{minipage}{\linewidth} 
\centering 
\includegraphics[height=19mm]{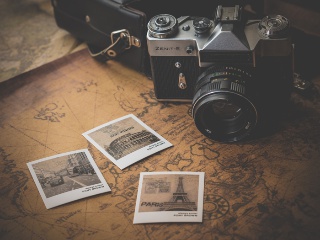}
\end{minipage}
\\
\vspace{3mm}
\\
\begin{minipage}{\linewidth} 
\centering 
\scriptsize
\textbf{Style 3}
\end{minipage}
\\
\begin{minipage}{\linewidth}
\centering 
\includegraphics[width=\linewidth]{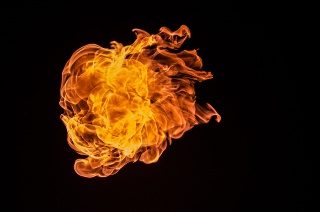}
\end{minipage}
\end{minipage}
\caption{\label{fig:photorealistic_style_transfer}Photorealistic style transfer results obtained by~\cite{luan17} (second column) and our network without (third column) and with graph filtering (fourth column), for $3$ different content images (first column) and corresponding style (last column). Zoom in different regions are presented below each result for better visualisation of fine details.}
\end{figure}

Another fundamental part of the method in \cite{luan17} is to match parts of the content and style images that have the same semantic, \eg, sky to sky, buildings to buildings, roads to roads, etc. This matching can be achieved by using masks in the style loss $\set{L}_s$. Let $\ma{M}^c \in \Rbb^{n \times n}$ and $\ma{M}^s \in \Rbb^{n' \times n'}$ denote the masks applying on $\Sig_c$ and $\Sig_s$ to match similar semantic regions. The masked style loss satisfies
\begin{align*}
\set{L}_s(\Sig) 
= 
\frac{1}{\abs{\mathcal{I}_s}} \; \sum_{\ell \in \mathcal{I}_s}  \frac{1}{c_\ell^2} \, \norm{ \frac{1}{{\rm Tr}(\ma{M}^c_\ell)} \, (\ma{M}^c_\ell \ma{F}_\ell)^\adjoint (\ma{M}^c_\ell \ma{F}_\ell) - \frac{1}{{\rm Tr}(\ma{M}^s_\ell)} \, (\ma{M}^s_\ell \ma{S}_\ell)^\adjoint (\ma{M}^s_\ell \ma{S}_\ell)}_{\rm F}^2,
\end{align*}
where $\ma{M}^c_\ell$ and $\ma{M}^s_\ell$ are the masks propagated at the $\ell^\th$ layer of the VGG-19. The partial derivative (\ref{eq:partial_derivative_style}) becomes
\begin{align}
\frac{\partial\set{L}_s}{\partial \ma{F}_\ell} \propto (\ma{M}^c_\ell \ma{F}_\ell) \cdot \left[ \frac{1}{{\rm Tr}(\ma{M}^c_\ell)} \, (\ma{M}^c_\ell \ma{F}_\ell)^\adjoint (\ma{M}^c_\ell \ma{F}_\ell) - \frac{1}{{\rm Tr}(\ma{M}^s_\ell)} \, (\ma{M}^s_\ell \ma{S}_\ell)^\adjoint (\ma{M}^c_\ell \ma{S}_\ell) \right].
\end{align}
The \emph{style correction filter} is thus computed by taking into account the semantic regions of interest only. By copying these modifications in \eqref{eq:estimated_pd}, we obtain
\begin{align}
\label{eq:masked_estimated_pd}
h_{\ell}^f (\Sig^{(t)}) \cdot \left[ \frac{1}{{\rm Tr}(\ma{M}^c_\ell)} \, \Big(\ma{M}^c_\ell h_{\ell}^f (\Sig^{(t)}) \Big)^\adjoint \, \Big(\ma{M}^c_\ell h_{\ell}^f (\Sig^{(t)}) \Big)
\; - \; \ma{H}_{\ell, t}^s \right].
\end{align}
Note that $\ma{H}_{\ell, t}^s$ encodes the style of a single semantic region by construction of the training procedure, hence the absence of modification of this matrix at runtime. Let us highlight that we did not multiply $h_{\ell}^f (\Sig^{(t)})$ with $\ma{M}^c_\ell$ on the left-hand side of \eqref{eq:masked_estimated_pd}. We noticed that this masking creates artefacts at the boundary of the mask. Instead, we mask $\Sig^{(4)}$ to keep the effect of the style transfer on the semantic region of interest only. Masking $h_{\ell}^f (\Sig^{(t)})$ to compute the \emph{style correction filter} is however necessary as, otherwise, one takes into account the whole image in the amount of correction instead of just the semantic region to correct. Let us highlight that to recombine the different stylised semantic regions, we do not apply binary masks on $\Sig^{(4)}$ but alpha mattes computed using the method of \emph{Levin et al.}~\cite{levin08} and user-defined strokes.

\section{Style Transfer Experiments}
\begin{figure}[t]
\centering
\includegraphics[width=0.19\linewidth]{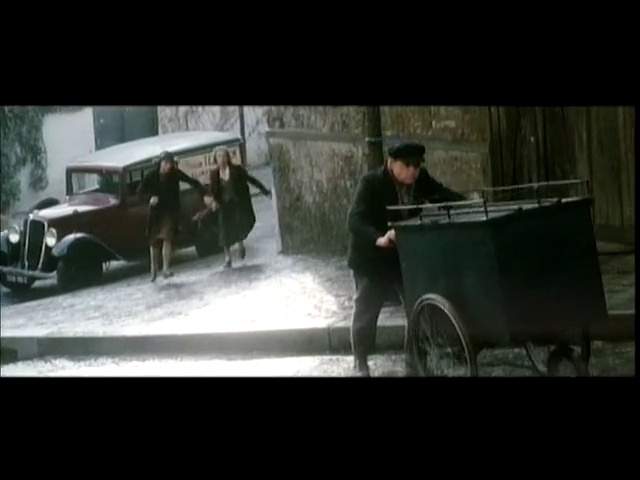}
\includegraphics[width=0.19\linewidth]{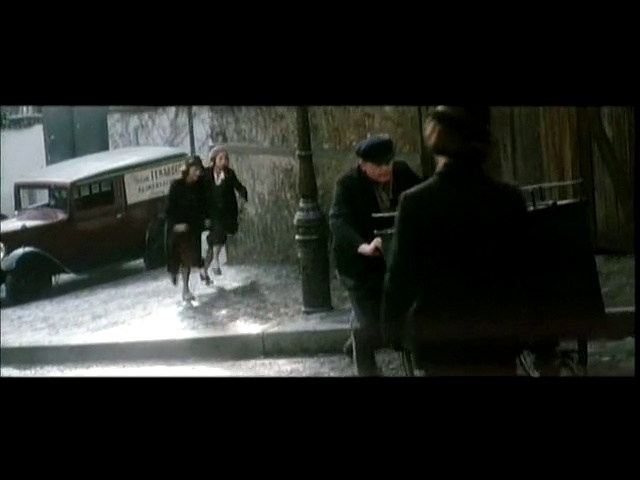}
\includegraphics[width=0.19\linewidth]{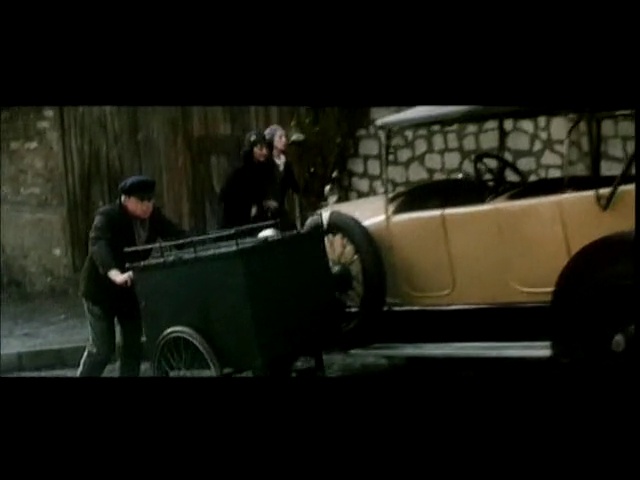}
\includegraphics[width=0.19\linewidth]{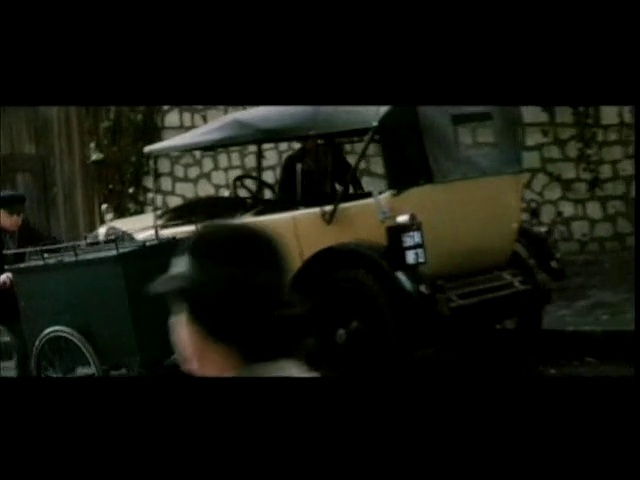}
\\
\includegraphics[width=0.19\linewidth]{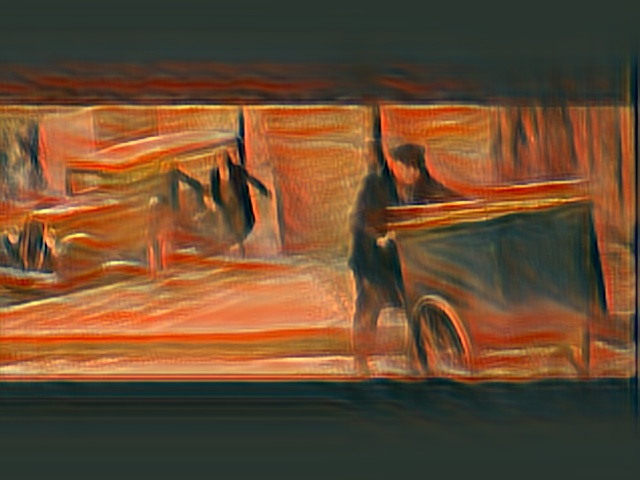}
\includegraphics[width=0.19\linewidth]{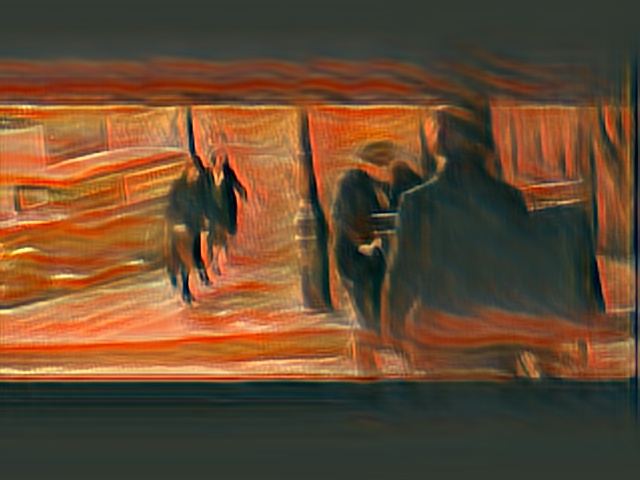}
\includegraphics[width=0.19\linewidth]{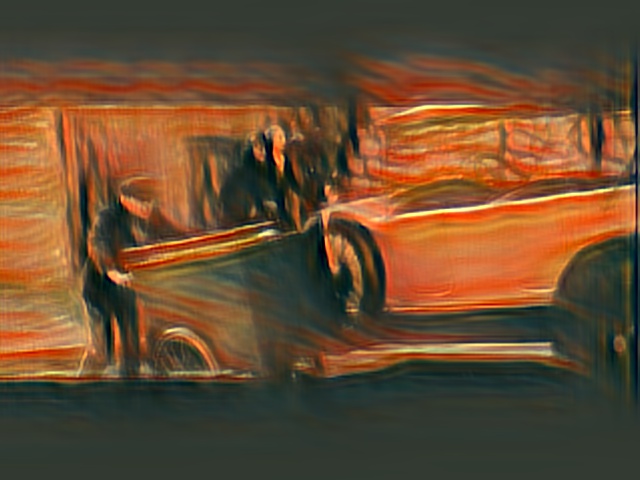}
\includegraphics[width=0.19\linewidth]{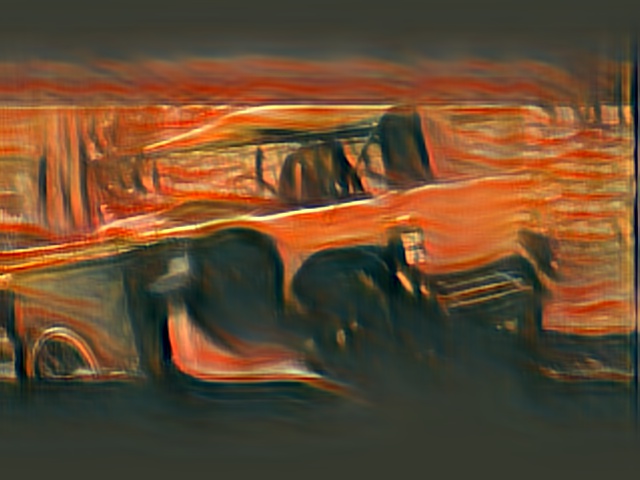}
\\
\includegraphics[width=0.19\linewidth]{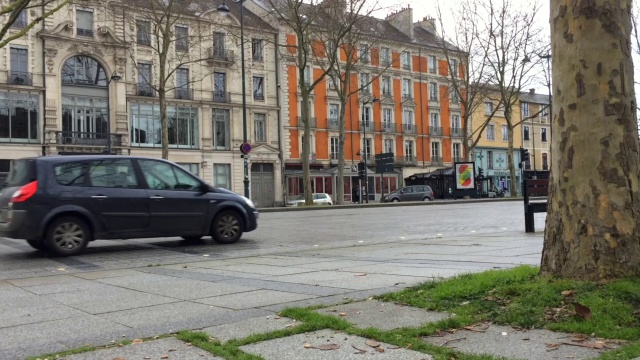}
\includegraphics[width=0.19\linewidth]{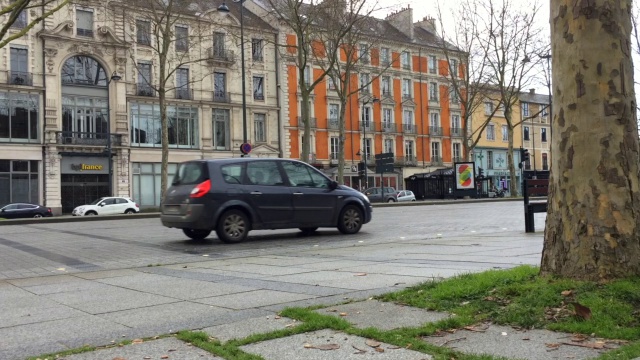}
\includegraphics[width=0.19\linewidth]{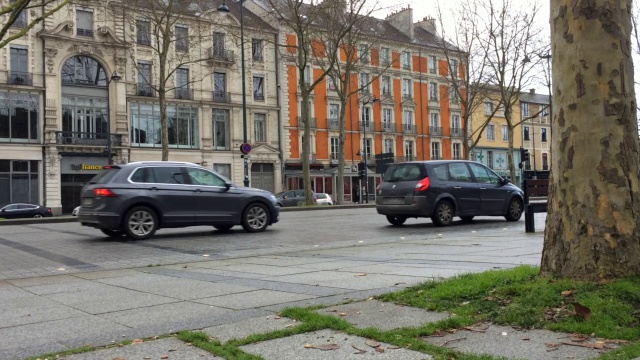}
\includegraphics[width=0.19\linewidth]{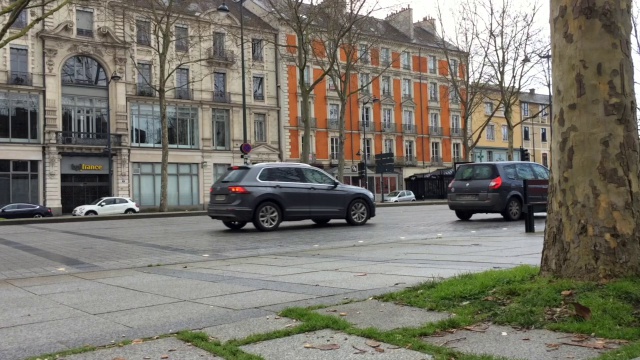}
\\
\includegraphics[width=0.19\linewidth]{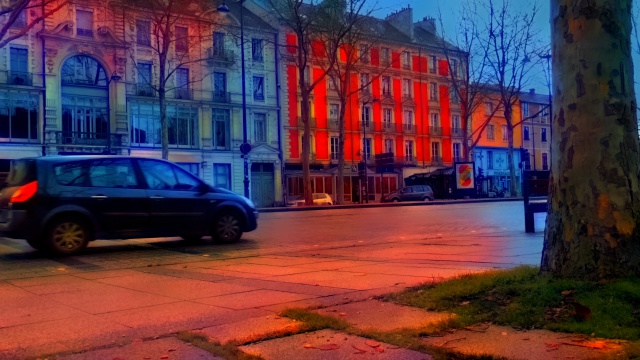}
\includegraphics[width=0.19\linewidth]{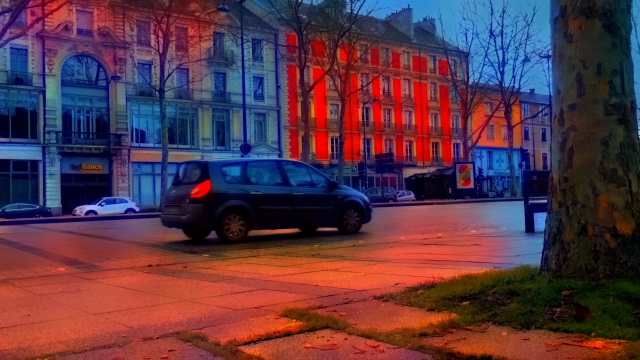}
\includegraphics[width=0.19\linewidth]{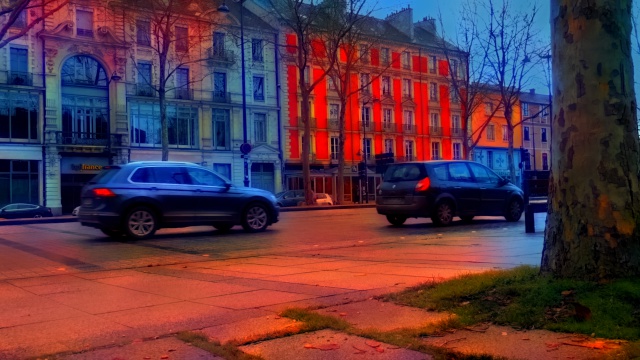}
\includegraphics[width=0.19\linewidth]{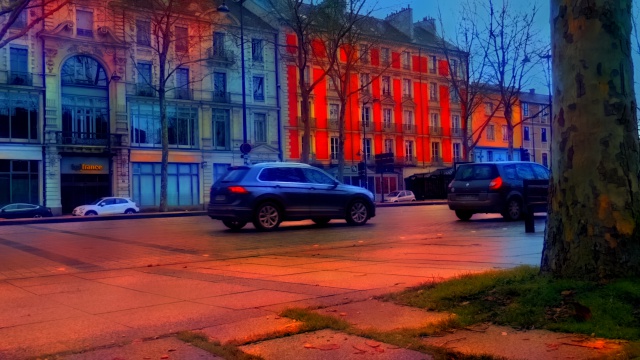}
\\
\caption{\label{fig:video_style_transfer}Style transfer results on two video sequences (first and third rows) for artistic style $1$ of Fig.~\ref{fig:artistic_style_transfer} (second row) and photorealistic style 1 of Fig.~\ref{fig:photorealistic_style_transfer} (last row).}
\end{figure}
%

\textbf{Fast artistic style transfer.} 
We present in Fig.~\ref{fig:artistic_style_transfer} artistic style transfer results. We first show results obtained using the method of Gatys\emph{ et al}.~\cite{gatys16} by minimising \eqref{eq:training_cost} using the L-BFGS algorithm. We use random noise as initialisation of this algorithm, as originally done in \cite{gatys16}. We also initialise the L-BFGS algorithm with $\Sig_c$ as our network is trained to minimise \eqref{eq:training_cost} starting form this initial image. Finally, we also present in the same figure the corresponding results obtained using our trained network. We notice that we obtain images qualitatively similar to those obtained by \cite{gatys16}. As a quantitative result, we show in Fig.~\ref{fig:fb_network_graph} the evolution of \eqref{eq:training_cost} as a function of the number of iterations of the L-BFGS algorithm and the value of \eqref{eq:training_cost} reached by our network. The curves show values averaged over $5$ test images using for $2$ different styles. Our method achieves a loss comparable to around $40$ iterations of L-BFGS when initialized with noise. Similar results are obtained in \cite{johnson2016perceptual}. This shows that our network achieves results close to state-of-the-art methods for fast artistic style transfer.

\noindent\textbf{Fast photorealistic style transfer.}
We present in Fig.~\ref{fig:photorealistic_style_transfer} style transfer results for styles that are photographs. The results are presented both with and without multiscale graph filtering at runtime. For comparison, we also present the results\footnote{Downloaded at \url{https://github.com/luanfujun/deep-photo-styletransfer}} obtained by Luan \emph{et al}.~\cite{luan17}. First, we notice that graph filtering permits to correct geometric distortions and improve the photorealism. Second, we claim that we achieve results of similar overall quality as Luan \emph{et al}.~\cite{luan17}. We notice however that we preserve better fine details. This is more visible in close-ups below each image, especially in those containing text.

\noindent\textbf{Video style transfer.} We present in Fig.~\ref{fig:video_style_transfer} artistic and photorealistic style transfer results on four frames of two video sequences. Even though each frame is processed individually, we remark that the results are relatively consistent temporally. The processing time was $0.10$~s and $8.4$~s per frame per style for the artistic -- on frames of $640 \times 480$ pixels -- and photorealistic -- on frames of $640 \times 360$ pixels -- style transfers, respectively.\footnote{The computation time does not include the computation of $\Lap$ nor the final guided filtering step, and was measured on an Nvidia Tesla M60 Maxwell GPU.} The increase in computational time is due to graph filtering. Even though graph filtering is computationally efficient, we still have to filter all the feature maps --~$428$ in total~-- at all the scales of the backward maps. One could reduce the computational time, by, \eg, filtering only $g_t(\cdot)$. We noticed however that this usually requires fine tuning of the graph filtering parameters for each pair of content and style images to achieve satisfactory results.

\section{Discussion and Conclusion}
\label{sec:conclusion}
\begin{figure}[t]
\centering
\begin{minipage}{0.19\linewidth}\centering \scriptsize $\Sig_c$ \end{minipage}
\begin{minipage}{0.19\linewidth}\centering \scriptsize \cite{luan17} \end{minipage}
\begin{minipage}{0.19\linewidth}\centering \scriptsize \cite{li18}\end{minipage}
\begin{minipage}{0.19\linewidth}\centering \scriptsize Ours \end{minipage}
\begin{minipage}{0.19\linewidth}\centering \scriptsize $\Sig_s$ \end{minipage}
\\
\begin{minipage}{0.19\linewidth}
\includegraphics[width=\linewidth]{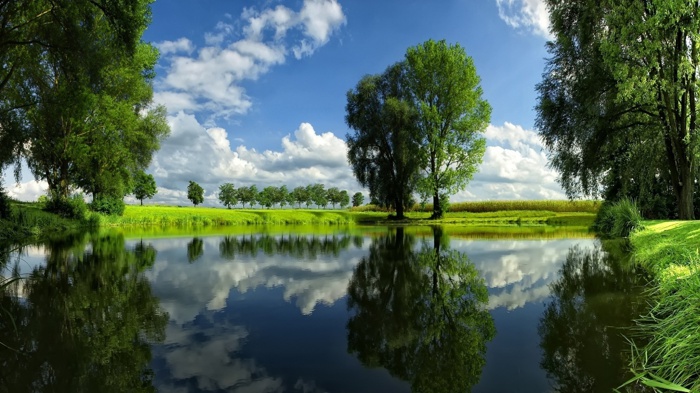}
\end{minipage}
\begin{minipage}{0.19\linewidth}
\includegraphics[width=\linewidth]{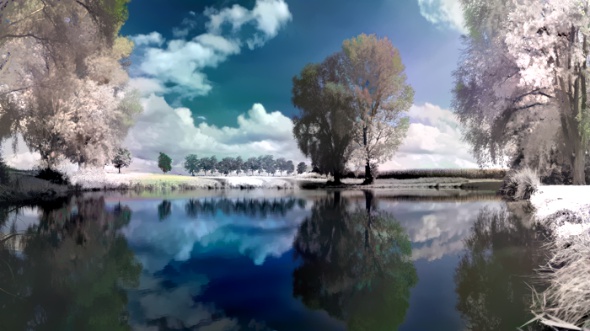}
\end{minipage}
\begin{minipage}{0.19\linewidth}
\includegraphics[width=\linewidth]{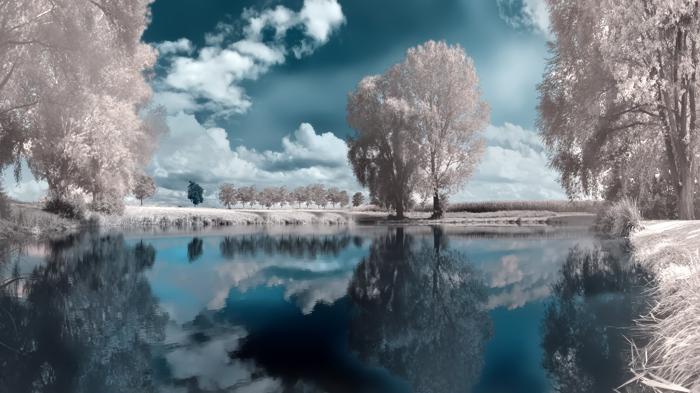}
\end{minipage}
\begin{minipage}{0.19\linewidth}
\includegraphics[width=\linewidth]{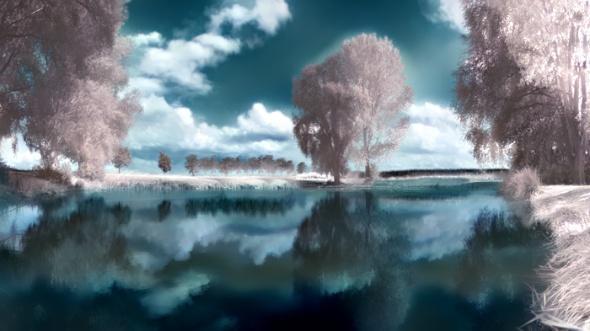}
\end{minipage}
\begin{minipage}{0.19\linewidth}
\centering
\includegraphics[height=16mm]{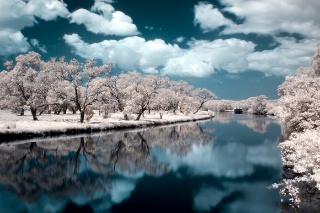}
\end{minipage}
\caption{\label{fig:comparison}Style transfer results obtained by the optimisation method of \cite{luan17}, the fast method of \cite{li18} and us.}
\end{figure}

A technical report proposing a fast method for photorealistic style transfer was made public \cite{li18} while writing this report. The authors adapt the deep network of \cite{li17} used for fast artistic style transfer: the deep decoders trained to invert different levels of the VGG-19 in \cite{li17} now use unpooling layers instead of upsampling layers. A non-photorealistic stylised image is then obtained exactly as in \cite{li17} but using these new decoders. To achieve a photorealistic result, this intermediate image is post-processed by solving a quadratic problem involving a graph regularisation term with the matting Laplacian of \cite{levin08}. Our method is fundamentally different. First, the network architecture is derived by unrolling the gradient descent algorithm, which is a novel approach for style transfer. Second, photorealism is obtained by modifying the network at runtime, not by post-processing. Third, our network contains only $281\,795$ parameters once a style is fixed, while the networks in \cite{li17} contains more than $7$ million parameters, including about $3.5$ million pre-trained parameters of the VGG-19. We remark however that the method of \cite{li18} does not need retraining for each new style whereas ours does. We present in Fig.~\ref{fig:comparison} results obtained with this method and ours. We notice that we achieve similar qualities to \cite{li18}, whereas the color palette of the result obtained by \cite{luan17} still contains colors of the content image. Finally, we also remark that no results on videos are presented in \cite{li18}.
 
We proposed a new deep neural network architecture for fast artistic style transfer which has a key advantage: it can be restructured at runtime to incorporate, \eg, the photorealistic prior proposed by Luan \emph{et al.}~\cite{luan17}. This property is made possible because our network architecture is derived by unrolling the gradient descent algorithm used to minimise Gatys \emph{et al.}'s style transfer loss: modification of this loss implies modifications of the gradient descent algorithm which can be passed on our network architecture. Finally, let us highlight that, beyond the style transfer problem, this principle could be used to derive flexible deep architectures architectures to solve other problems formulated as a minimisation of a loss function.

%
\titleformat{\section}[hang]{\color{blue1}\large\bfseries}{\thesection}{0mm}{}[]
\bibliographystyle{IEEEtran}
\bibliography{deepsolver}

\begin{thebibliography}{10}
\providecommand{\url}[1]{#1}
\csname url@samestyle\endcsname
\providecommand{\newblock}{\relax}
\providecommand{\bibinfo}[2]{#2}
\providecommand{\BIBentrySTDinterwordspacing}{\spaceskip=0pt\relax}
\providecommand{\BIBentryALTinterwordstretchfactor}{4}
\providecommand{\BIBentryALTinterwordspacing}{\spaceskip=\fontdimen2\font plus
\BIBentryALTinterwordstretchfactor\fontdimen3\font minus
  \fontdimen4\font\relax}
\providecommand{\BIBforeignlanguage}[2]{{%
\expandafter\ifx\csname l@#1\endcsname\relax
\typeout{** WARNING: IEEEtran.bst: No hyphenation pattern has been}%
\typeout{** loaded for the language `#1'. Using the pattern for}%
\typeout{** the default language instead.}%
\else
\language=\csname l@#1\endcsname
\fi
#2}}
\providecommand{\BIBdecl}{\relax}
\BIBdecl

\bibitem{gatys16}
L.~A. Gatys, A.~S. Ecker, and M.~Bethge, ``Image style transfer using
  convolutional neural networks,'' in \emph{CVPR}, 2016.

\bibitem{luan17}
F.~Luan, S.~Paris, E.~Shechtman, and K.~Bala, ``Deep photo style transfer,'' in
  \emph{IEEE Conference on Computer Vision and Pattern Recognition (CVPR)},
  2017, pp. 4990--4998.

\bibitem{gatys15}
L.~A. Gatys, A.~S. Ecker, and Bet, ``Texture synthesis using convolutional
  neural networks,'' in \emph{NIPS}, 2015.

\bibitem{johnson2016perceptual}
J.~Johnson, A.~Alahi, and L.~Fei-Fei, ``Perceptual losses for real-time style
  transfer and super-resolution,'' in \emph{ECCV}, 2016.

\bibitem{ulyanov2016texture}
D.~Ulyanov, V.~Lebedev, A.~Vedaldi, and V.~Lempitsky, ``Texture networks:
  Feed-forward synthesis of textures and stylized images,'' 2016.

\bibitem{chen17}
D.~Chen, L.~Yuan, J.~Liao, N.~Yu, and G.~Hua, ``Stylebank: An explicit
  representation for neural image style transfer,'' in \emph{CVPR}, 2017.

\bibitem{dumoulin17}
V.~Dumoulin, J.~Shlens, and M.~Kudlur, ``A learned representation for artistic
  style,'' in \emph{ICLR}, 2017.

\bibitem{li17}
Y.~Li, C.~Fang, J.~Yang, Z.~Wang, X.~Lu, and M.-H. Yang, ``Universal style
  transfer via feature transforms,'' in \emph{NIPS}, 2017.

\bibitem{huang17}
X.~Huang and S.~Belongie, ``Arbitrary style transfer in real-time with adaptive
  instance normalization,'' in \emph{ICCV}, 2017, pp. 1501--1510.

\bibitem{ghiasi17}
G.~Ghiasi, H.~Lee, M.~Kudlur, V.~Dumoulin, and J.~Shlens, ``Exploring the
  structure of a real-time, arbitrary neural artistic stylization network,'' in
  \emph{BMVC}, 2017.

\bibitem{li18}
Y.~Li, M.-Y. Liu, X.~Li, M.-H. Yang, and J.~Kautz, ``A closed-form solution to
  photorealistic image stylization,'' \emph{arXiv:1802.06474}, 2018.

\bibitem{reinhard2001color}
E.~Reinhard, M.~Adhikhmin, B.~Gooch, and P.~Shirley, ``Color transfer between
  images,'' \emph{IEEE Computer graphics and applications}, vol.~21, no.~5, pp.
  34--41, 2001.

\bibitem{efros2001image}
A.~A. Efros and W.~T. Freeman, ``Image quilting for texture synthesis and
  transfer,'' in \emph{SIGGRAPH}, 2001.

\bibitem{hertzmann2001image}
A.~Hertzmann, C.~E. Jacobs, N.~Oliver, B.~Curless, and D.~H. Salesin, ``Image
  analogies,'' in \emph{SIGGRAPH}, 2001.

\bibitem{he17}
M.~He, J.~Liao, L.~Yuan, and P.~V. Sander, ``Neural color transfer between
  images,'' \emph{arXiv:1710.00756}, 2017.

\bibitem{gregor2010learning}
K.~Gregor and Y.~LeCun, ``Learning fast approximations of sparse coding,'' in
  \emph{ICML}, 2010.

\bibitem{beck2009fast}
A.~Beck and M.~Teboulle, ``A fast iterative shrinkage-thresholding algorithm
  for linear inverse problems,'' \emph{SIAM journal on imaging sciences},
  vol.~2, no.~1, pp. 183--202, 2009.

\bibitem{mousavi2017learning}
A.~Mousavi and R.~G. Baraniuk, ``Learning to invert: Signal recovery via deep
  convolutional networks,'' in \emph{ICASSP}, 2017.

\bibitem{metzler2017learned}
C.~Metzler, A.~Mousavi, and R.~Baraniuk, ``Learned d-amp: Principled neural
  network based compressive image recovery,'' in \emph{NIPS}, 2017.

\bibitem{borgerding2017amp}
M.~Borgerding, P.~Schniter, and S.~Rangan, ``Amp-inspired deep networks for
  sparse linear inverse problems,'' \emph{IEEE Trans. on Signal Processing},
  2017.

\bibitem{zhang2017ista}
J.~Zhang and B.~Ghanem, ``Ista-net: Iterative shrinkage-thresholding algorithm
  inspired deep network for image compressive sensing,'' \emph{arXiv preprint
  arXiv:1706.07929}, 2017.

\bibitem{wang2016proximal}
S.~Wang, S.~Fidler, and R.~Urtasun, ``Proximal deep structured models,'' in
  \emph{NIPS}, 2016.

\bibitem{liu2017proximal}
R.~Liu, X.~Fan, S.~Cheng, X.~Wang, and Z.~Luo, ``Proximal alternating direction
  network: A globally converged deep unrolling framework,'' \emph{arXiv
  preprint arXiv:1711.07653}, 2017.

\bibitem{chen2017trainable}
Y.~Chen and T.~Pock, ``Trainable nonlinear reaction diffusion: A flexible
  framework for fast and effective image restoration,'' \emph{IEEE transactions
  on pattern analysis and machine intelligence}, vol.~39, no.~6, pp.
  1256--1272, 2017.

\bibitem{tompson2016accelerating}
J.~Tompson, K.~Schlachter, P.~Sprechmann, and K.~Perlin, ``Accelerating
  eulerian fluid simulation with convolutional networks,'' \emph{arXiv preprint
  arXiv:1607.03597}, 2016.

\bibitem{tewari2017mofa}
A.~Tewari, M.~Zollh{\"o}fer, H.~Kim, P.~Garrido, F.~Bernard, P.~P{\'e}rez, and
  C.~Theobalt, ``Mofa: Model-based deep convolutional face autoencoder for
  unsupervised monocular reconstruction,'' in \emph{ICCV}, 2017.

\bibitem{simonyan14}
K.~Simonyan and A.~Zisserman, ``Very deep convolutional networks for
  large-scale image recognition,'' \emph{arXiv:1409.1556}, 2014.

\bibitem{he_resnet16}
K.~He, X.~Zhang, S.~Ren, and J.~Sun, ``Deep residual learning for image
  recognition,'' in \emph{CVPR}, 2016, pp. 770--778.

\bibitem{glorot10}
\BIBentryALTinterwordspacing
X.~Glorot and Y.~Bengio, ``Understanding the difficulty of training deep
  feedforward neural networks,'' in \emph{Proceedings of the Thirteenth
  International Conference on Artificial Intelligence and Statistics}, ser.
  Proceedings of Machine Learning Research, Y.~W. Teh and M.~Titterington,
  Eds., vol.~9.\hskip 1em plus 0.5em minus 0.4em\relax Chia Laguna Resort,
  Sardinia, Italy: PMLR, 13--15 May 2010, pp. 249--256. [Online]. Available:
  \url{http://proceedings.mlr.press/v9/glorot10a.html}
\BIBentrySTDinterwordspacing

\bibitem{kingma14}
D.~P. Kingma and J.~Ba, ``Adam: A method for stochastic optimization,''
  \emph{arXiv:1412.6980}, 2014.

\bibitem{lin14}
\BIBentryALTinterwordspacing
T.-Y. Lin, M.~Maire, S.~Belongie, J.~Hays, P.~Perona, D.~Ramanan, P.~Dollar,
  and L.~Zitnick, ``Microsoft coco: Common objects in context,'' in
  \emph{ECCV}.\hskip 1em plus 0.5em minus 0.4em\relax European Conference on
  Computer Vision, September 2014. [Online]. Available:
  \url{https://www.microsoft.com/en-us/research/publication/microsoft-coco-common-objects-in-context/}
\BIBentrySTDinterwordspacing

\bibitem{levin08}
A.~Levin, D.~Lischinski, and Y.~Weiss, ``A closed-form solution to natural
  image matting,'' \emph{IEEE Trans. Pattern Anal. Mach. Intell.}, vol.~30,
  no.~2, pp. 228--242, 2008.

\bibitem{shuman_SPMAG2013}
D.~Shuman, S.~Narang, P.~Frossard, A.~Ortega, and P.~Vandergheynst, ``The
  emerging field of signal processing on graphs: Extending high-dimensional
  data analysis to networks and other irregular domains,'' \emph{IEEE Signal
  Processing Magazine}, vol.~30, no.~3, pp. 83--98, 2013.

\bibitem{hammond11}
D.~K. Hammond, P.~Vandergheynst, and R.~Gribonval, ``Wavelets on graphs via
  spectral graph theory,'' \emph{Appl. Comput. Harmon. Anal.}, vol.~30, no.~2,
  pp. 129--150, 2011.

\bibitem{napoli13}
E.~D. Napoli, E.~Polizzo, and Y.~Saad, ``Efficient estimation of eigenvalue
  counts in an interval,'' \emph{Numerical Linear Algebra with Applications},
  vol.~23, no.~4, pp. 674--692, 2016.

\bibitem{defferrard16}
M.~Defferrard, X.~Bresson, and P.~Vandergheynst, ``Convolutional neural
  networks on graphs with fast localized spectral filtering,'' in \emph{NIPS},
  2016.

\bibitem{kipf16}
T.~N. Kipf and M.~Welling, ``Semi-supervised classification with graph
  convolutional networks,'' \emph{arXiv:1609.02907}, 2016.

\end{thebibliography}

%
\titleformat{\section}[hang]{\color{blue1}\large\bfseries\centering}{Appendix \thesection\ - }{0mm}{}[]
\appendix
%
\section{Varying the Intensity of the Transfer}
\label{sec:vary_intensity}

The intensity of the transfer can be modified by adjusting the parameter $\lambda_s$ in 
\begin{align}
\set{L}(\Sig) = \lambda_c \, \set{L}_c(\Sig, \Sig_c) + \lambda_s \, \set{L}_s(\Sig, \Sig_s).
\end{align}
If one multiplies $\lambda_s$ by $\alpha > 0$, and keeps the stepsize $\mu$ constant, the gradient descent algorithm becomes
\begin{align} 
\Sig^{(t+1)} = \Sig^{(t)} - \mu \left[ \lambda_c \, \nabla\set{L}_c(\Sig^{(t)}) + \alpha \lambda_s \, \nabla\set{L}_s(\Sig^{(t)}) \right].
\end{align}
This suggests that the intensity of the transfer can be adjusted at runtime as follows
\begin{align}
\label{eq:varying_intensity}
\Sig^{(t+1)} = \Sig^{(t)} - \alpha \; g_{t} \left( \Sig^{(t)} \right),
\end{align}
in our network.

\begin{figure}
\centering
\begin{minipage}{0.19\linewidth}\centering $\alpha=0.6$ \end{minipage}
\begin{minipage}{0.19\linewidth}\centering $\alpha=0.8$ \end{minipage}
\begin{minipage}{0.19\linewidth}\centering $\alpha=1.0$ \end{minipage}
\begin{minipage}{0.19\linewidth}\centering $\alpha=1.2$ \end{minipage}
\\
\begin{minipage}{0.19\linewidth}
\includegraphics[width=\linewidth]{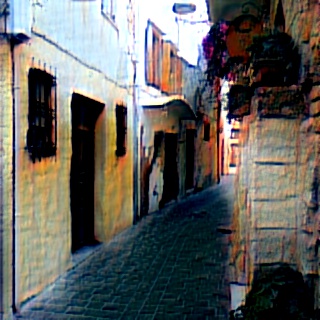}
\end{minipage}
\begin{minipage}{0.19\linewidth}
\includegraphics[width=\linewidth]{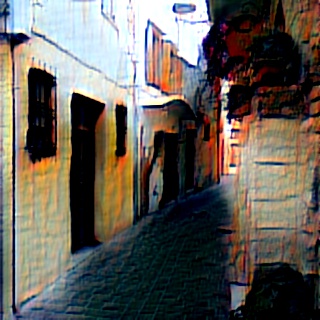}
\end{minipage}
\begin{minipage}{0.19\linewidth}
\includegraphics[width=\linewidth]{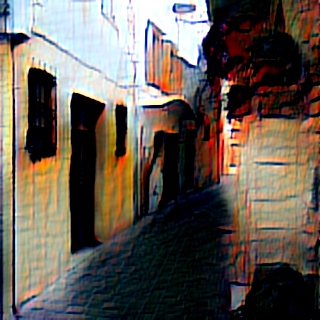}
\end{minipage}
\begin{minipage}{0.19\linewidth}
\includegraphics[width=\linewidth]{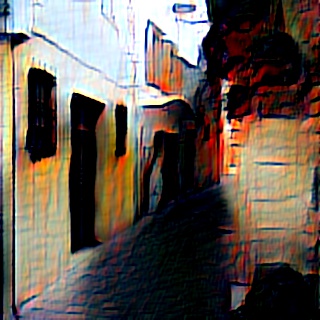}
\end{minipage}
\\
\begin{minipage}{0.19\linewidth}
\includegraphics[width=\linewidth]{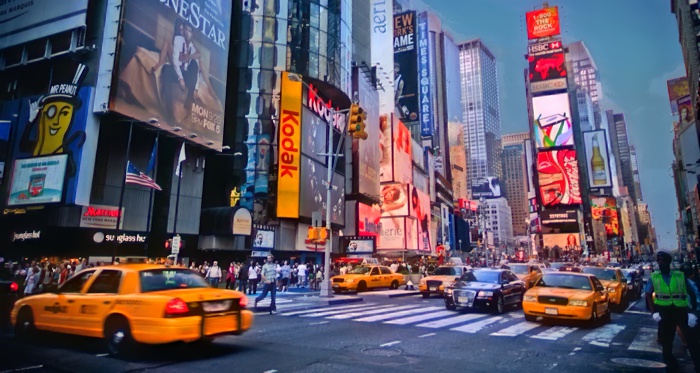}
\end{minipage}
\begin{minipage}{0.19\linewidth}
\includegraphics[width=\linewidth]{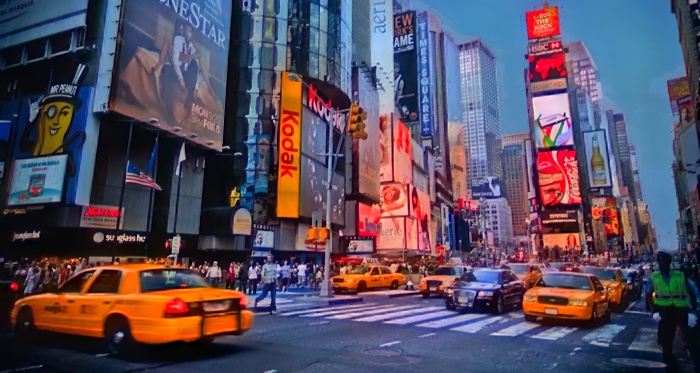}
\end{minipage}
\begin{minipage}{0.19\linewidth}
\includegraphics[width=\linewidth]{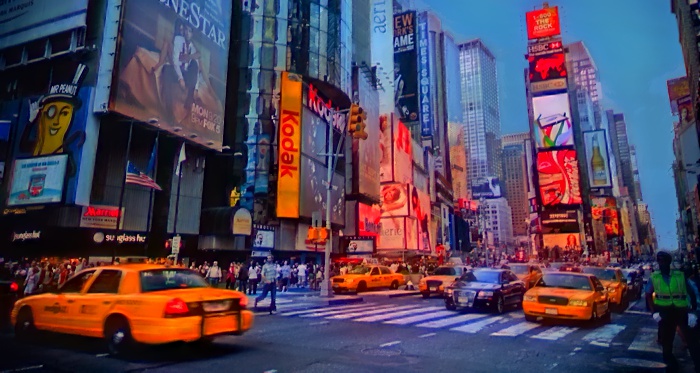}
\end{minipage}
\begin{minipage}{0.19\linewidth}
\includegraphics[width=\linewidth]{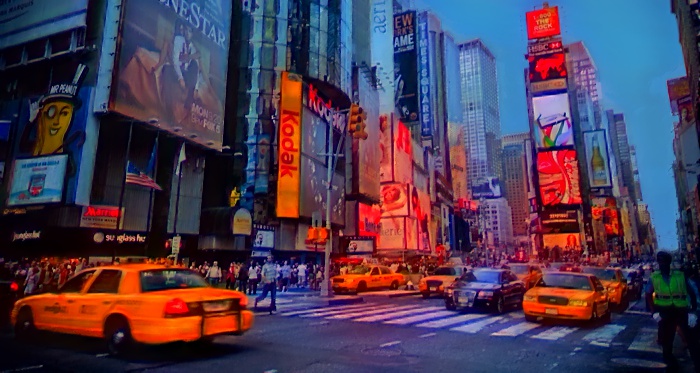}
\end{minipage}
\vspace{2mm}
\\
\begin{minipage}{0.24\linewidth} 
\centering
\scriptsize
\textbf{Content 1}
\end{minipage}
\begin{minipage}{0.24\linewidth} 
\centering
\scriptsize
\textbf{Style 1}\\
\emph{A Muse},\\
P.~Picasso,
1935.
\vspace{1mm}
\end{minipage}
\begin{minipage}{0.24\linewidth} 
\centering
\scriptsize
\textbf{Content 2}
\end{minipage}
\begin{minipage}{0.24\linewidth} 
\centering
\scriptsize
\textbf{Style 2}
\end{minipage}
\\
\begin{minipage}{0.24\linewidth} 
\centering
\includegraphics[height=24mm]{inputs/chania_a.jpg}
\end{minipage}
\begin{minipage}{0.24\linewidth} 
\centering
\includegraphics[height=24mm]{styles/tar70_i.jpg}
\end{minipage}
\begin{minipage}{0.24\linewidth} 
\includegraphics[width=\linewidth]{inputs/in10_i}
\end{minipage}
\begin{minipage}{0.24\linewidth} 
\includegraphics[width=\linewidth]{styles/tar10_i}
\end{minipage}
\\
\caption{\label{fig:varying_intensity}Adjusting the intensity of style transfer at runtime by varying $\alpha$ in \eqref{eq:varying_intensity} for an artistic result (top, with Content \& Style 1) or a photorealistic  result (bottom, with Content \& Style 2).}
\end{figure}

We show the effect of varying $\alpha$ in Fig.~\ref{fig:varying_intensity}. Results are presented for artistic and photorealistic style transfers. These results show that we still have a degree of freedom to adjust the intensity of the style, even though the network was trained with $\alpha=1$.

Let us highlight that all the results presented in this work are generated with $\alpha=1$ in absence of graph filtering and with $\alpha=1.2$ in presence of graph filtering in the network. This permits us to correct a loss in color saturation that we observe in most results when graph filtering is introduced in the network.

%

%
\section{Effect of the Guided Filter}
\label{sec:guided_filter}

As in \cite{luan17} and \cite{li18}, we authorise ourselves to post-process the obtained stylized images with a guided filter (using the content image as guide). Let us note that we did not use the code provided by \cite{luan17} or \cite{li18} but used the OpenCV implementation of the guided filter. We discuss in this section the effect of this post-processing. In particular, we show that this post-processing permits us to remove remaining artefacts, such as compression artefacts, or enhances some details smoothed during style transfer. However, let us emphasize that the effect of the guided filter is limited to the improvement of these details and that the graph filtering introduced at runtime almost yields final results.

\begin{figure}
\centering
\begin{minipage}{0.74\linewidth}
\centering
\begin{minipage}{0.32\linewidth}\centering Content \end{minipage}
\begin{minipage}{0.32\linewidth}\centering No post-proc. \end{minipage}
\begin{minipage}{0.32\linewidth}\centering With post-proc. \end{minipage}
\\
\begin{minipage}{0.32\linewidth}
\includegraphics[width=\linewidth]{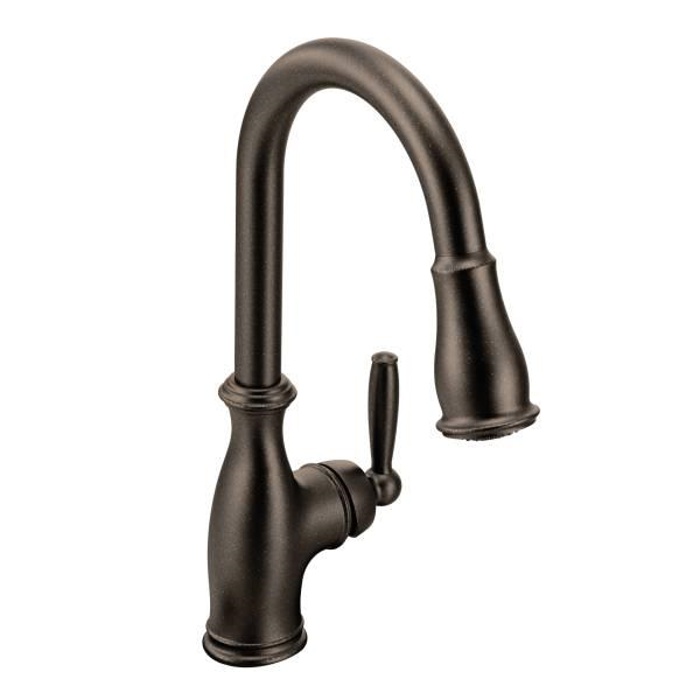}\\
\includegraphics[width=.31\linewidth]{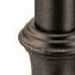}
\includegraphics[width=.31\linewidth]{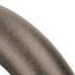}
\includegraphics[width=.31\linewidth]{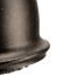}
\end{minipage}
\begin{minipage}{0.32\linewidth}
\includegraphics[width=\linewidth]{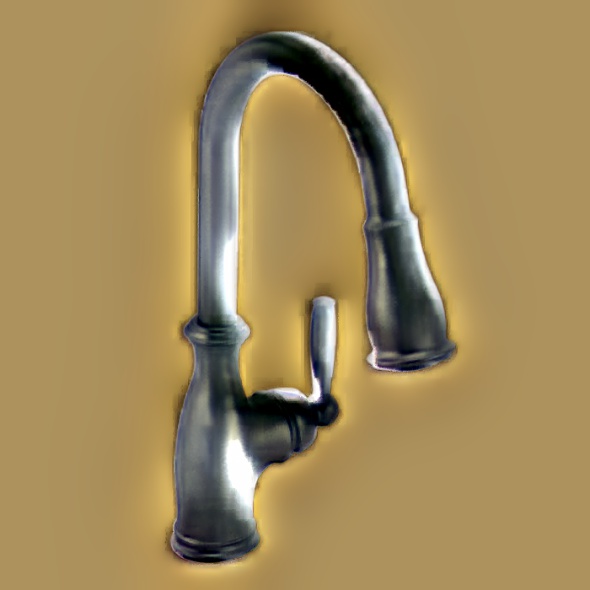}\\
\includegraphics[width=.31\linewidth]{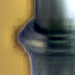}
\includegraphics[width=.31\linewidth]{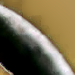}
\includegraphics[width=.31\linewidth]{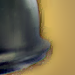}
\end{minipage}
\begin{minipage}{0.32\linewidth}
\includegraphics[width=\linewidth]{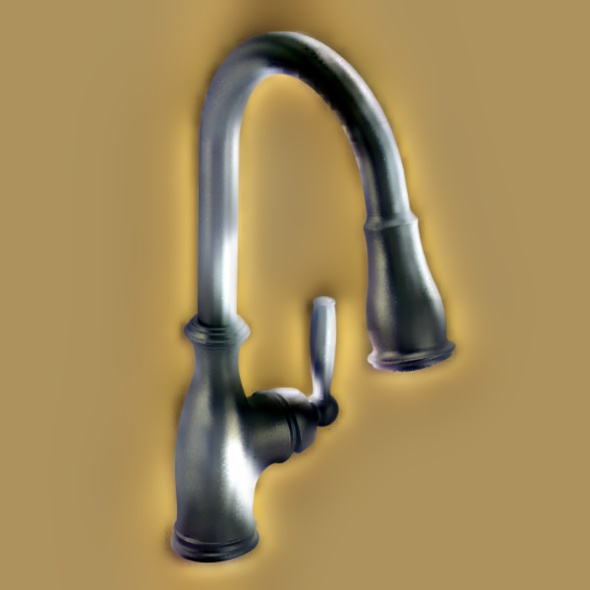}\\
\includegraphics[width=.31\linewidth]{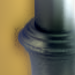}
\includegraphics[width=.31\linewidth]{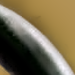}
\includegraphics[width=.31\linewidth]{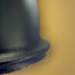}
\end{minipage}
\\
\begin{minipage}{0.32\linewidth}
\includegraphics[width=\linewidth]{inputs/in50_i}\\
\includegraphics[width=.31\linewidth]{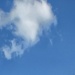}
\includegraphics[width=.31\linewidth]{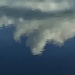}
\includegraphics[width=.31\linewidth]{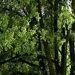}
\end{minipage}
\begin{minipage}{0.32\linewidth}
\includegraphics[width=\linewidth]{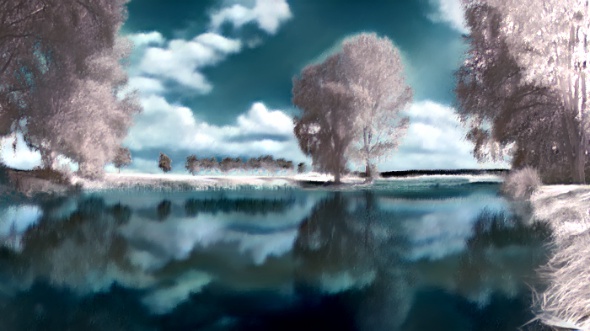}\\
\includegraphics[width=.31\linewidth]{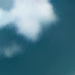}
\includegraphics[width=.31\linewidth]{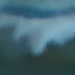}
\includegraphics[width=.31\linewidth]{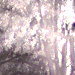}
\end{minipage}
\begin{minipage}{0.32\linewidth}
\includegraphics[width=\linewidth]{photos/res50_lap_post}\\
\includegraphics[width=.31\linewidth]{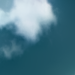}
\includegraphics[width=.31\linewidth]{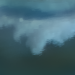}
\includegraphics[width=.31\linewidth]{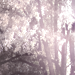}
\end{minipage}
\end{minipage}
\hfill
\vline
\hspace{0.5mm}
\begin{minipage}{0.23\linewidth}
\centering
\begin{minipage}{\linewidth}
\centering
\textbf{Style 1}
\includegraphics[width=\linewidth]{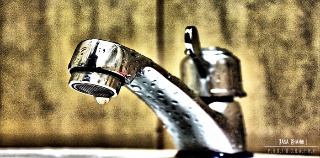}
\vspace{12mm}
\end{minipage}
\\
\begin{minipage}{\linewidth}
\centering
\textbf{Style 2}
\includegraphics[width=\linewidth]{styles/tar50_i}
\vspace{12mm}
\end{minipage}
\end{minipage}
\\
\caption{\label{fig:post_processing}Effect of post-processing with a guided filter. First column: Content images. Second: Results without guided filtering. Third: Results with guided filtering. Fourth: Style images. Please refer to Appendix~\ref{sec:guided_filter} for an interpretation of the results.}
\end{figure}

In the first example of Fig.~\ref{fig:post_processing}, we see that the guided filter removes artefacts that appear near the boundary of the object. These artefacts are most certainly due to image compression which perturbed the construction of the matting Laplacian, hence their presence after graph filtering. In the last example of Fig.~\ref{fig:post_processing}, we notice that the guided filter is able to sharpen some details and edges.

%

%
\section{Additional Style Transfer Results}
\label{sec:photo_results}

\begin{figure}
\centering
\begin{minipage}{0.24\linewidth}\centering Content \end{minipage}
\begin{minipage}{0.24\linewidth}\centering \cite{luan17} \end{minipage}
\begin{minipage}{0.24\linewidth}\centering Ours \end{minipage}
\begin{minipage}{0.24\linewidth}\centering Style \end{minipage}
\\
\begin{minipage}{0.24\linewidth}
\includegraphics[width=\linewidth]{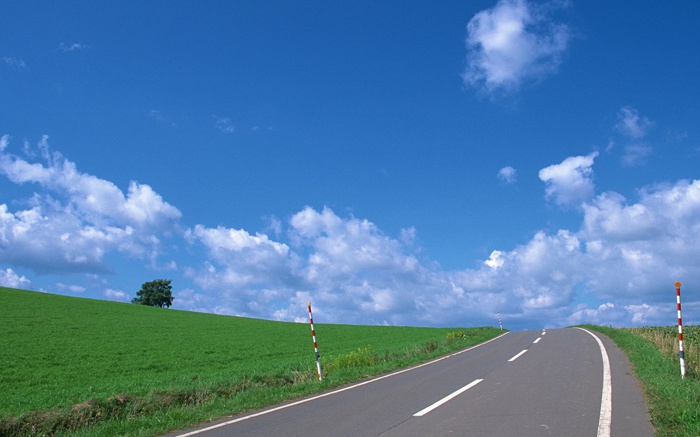}
\end{minipage}
\begin{minipage}{0.24\linewidth}
\includegraphics[width=\linewidth]{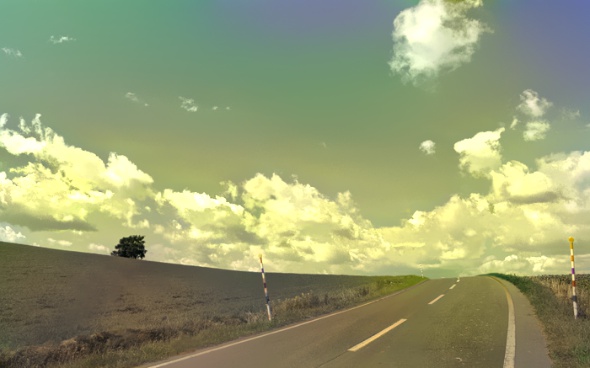}
\end{minipage}
\begin{minipage}{0.24\linewidth}
\includegraphics[width=\linewidth]{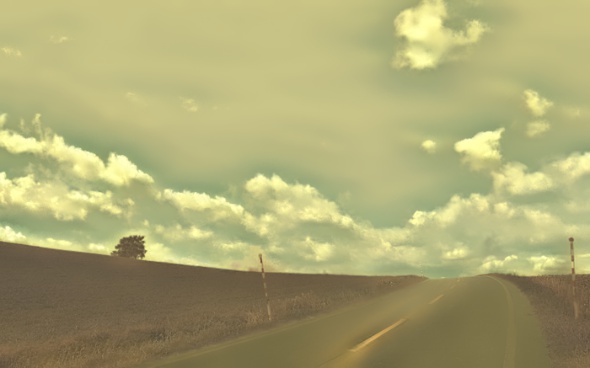}
\end{minipage}
\begin{minipage}{0.24\linewidth}
\includegraphics[width=\linewidth]{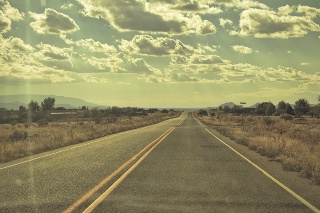}
\end{minipage}
\\
\begin{minipage}{0.24\linewidth}
\includegraphics[width=\linewidth]{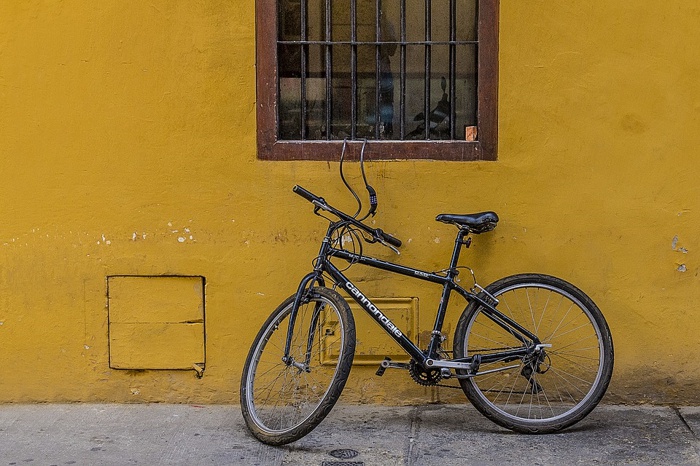}
\end{minipage}
\begin{minipage}{0.24\linewidth}
\includegraphics[width=\linewidth]{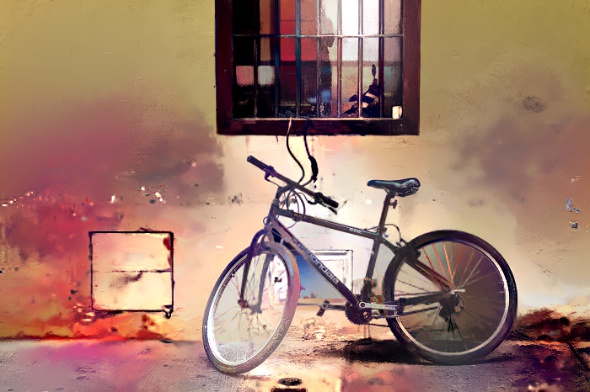}
\end{minipage}
\begin{minipage}{0.24\linewidth}
\includegraphics[width=\linewidth]{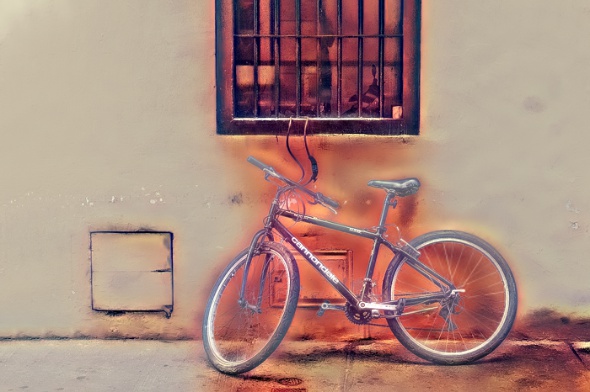}
\end{minipage}
\begin{minipage}{0.24\linewidth}
\includegraphics[width=\linewidth]{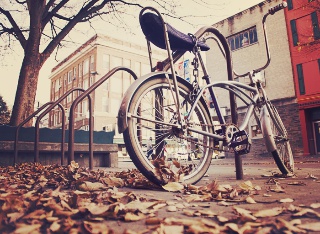}
\end{minipage}
\\
\begin{minipage}{0.24\linewidth}
\includegraphics[width=\linewidth]{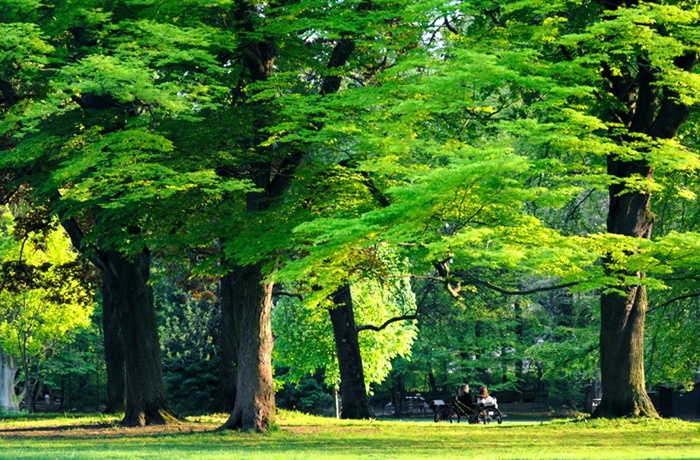}
\end{minipage}
\begin{minipage}{0.24\linewidth}
\includegraphics[width=\linewidth]{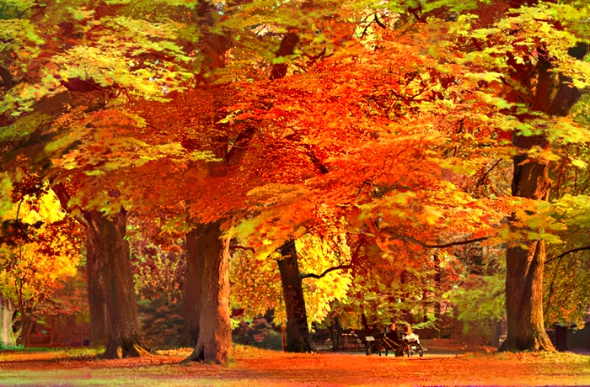}
\end{minipage}
\begin{minipage}{0.24\linewidth}
\includegraphics[width=\linewidth]{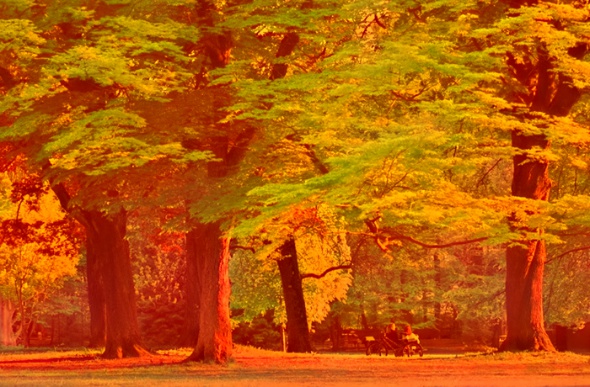}
\end{minipage}
\begin{minipage}{0.24\linewidth}
\centering
\includegraphics[width=\linewidth]{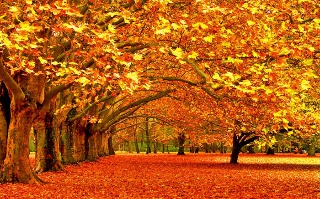}
\end{minipage}
\\
\begin{minipage}{0.24\linewidth}
\includegraphics[width=\linewidth]{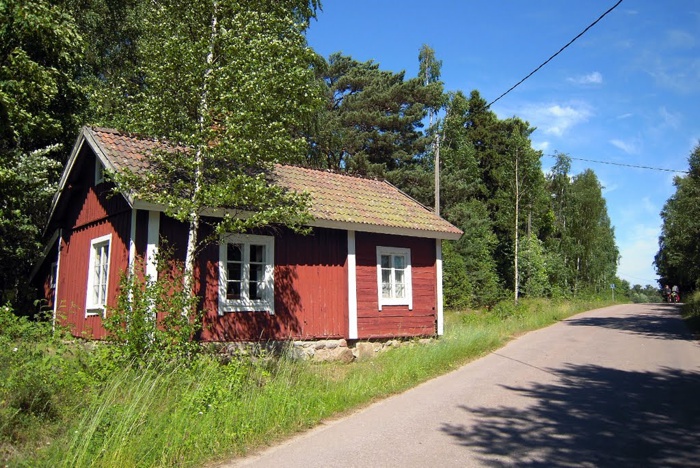}
\end{minipage}
\begin{minipage}{0.24\linewidth}
\includegraphics[width=\linewidth]{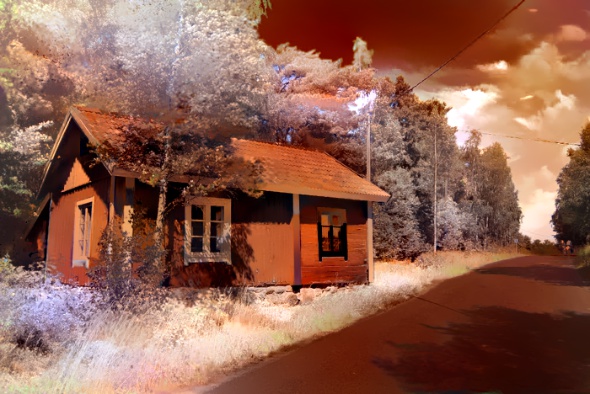}
\end{minipage}
\begin{minipage}{0.24\linewidth}
\includegraphics[width=\linewidth]{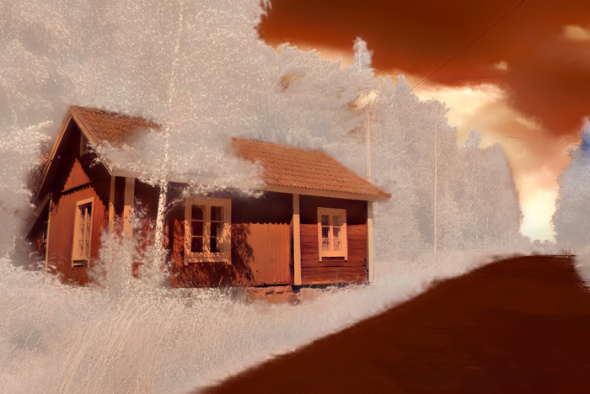}
\end{minipage}
\begin{minipage}{0.24\linewidth}
\centering
\includegraphics[width=\linewidth]{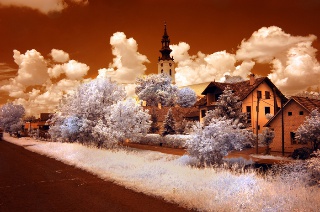}
\end{minipage}
\\
\begin{minipage}{0.24\linewidth}
\includegraphics[width=\linewidth]{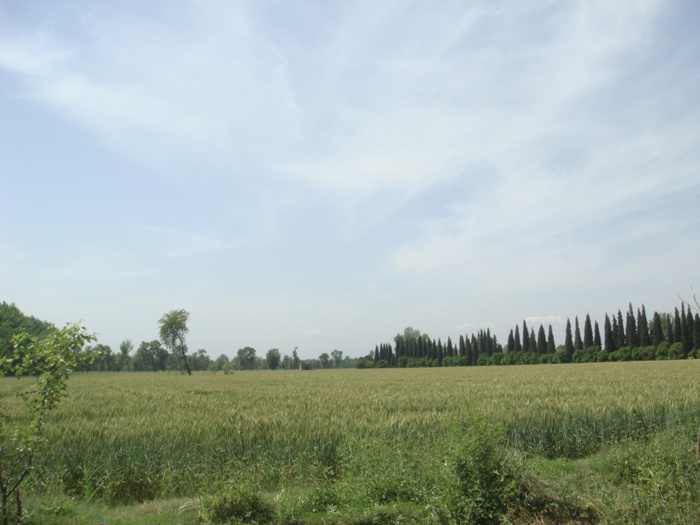}
\end{minipage}
\begin{minipage}{0.24\linewidth}
\includegraphics[width=\linewidth]{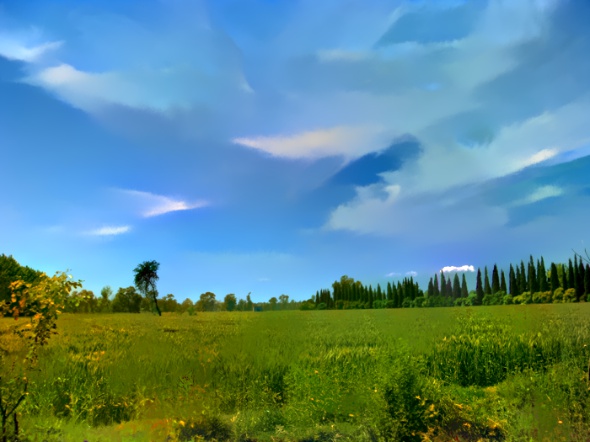}
\end{minipage}
\begin{minipage}{0.24\linewidth}
\includegraphics[width=\linewidth]{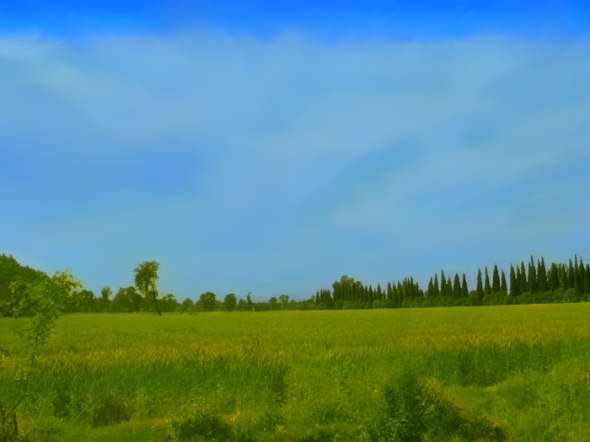}
\end{minipage}
\begin{minipage}{0.24\linewidth}
\includegraphics[width=\linewidth]{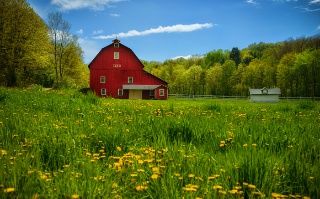}
\end{minipage}
\\
\caption{\label{fig:photorealistic_style_transfer_supp}Photorealistic style transfer results. First column: Content images. Second: Results obtained by \cite{luan17}. Third: Our results. Fourth: Style images. Please refer to Appendix~\ref{sec:photo_results} for comments about these results.}
\end{figure}
\begin{figure}
\centering
\begin{minipage}{0.75\linewidth}
\begin{minipage}{0.32\linewidth}\centering Content \end{minipage}
\begin{minipage}{0.32\linewidth}\centering \cite{gatys16} \end{minipage}
\begin{minipage}{0.32\linewidth}\centering Ours \end{minipage}
\\
\begin{minipage}{0.32\linewidth}
\includegraphics[width=\linewidth]{inputs/chania_a}
\end{minipage}
\begin{minipage}{0.32\linewidth}
\includegraphics[width=\linewidth]{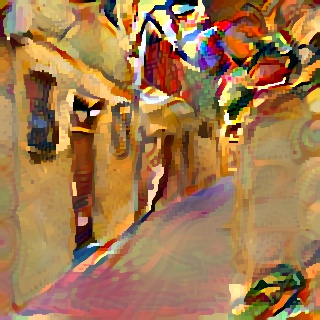}
\end{minipage}
\begin{minipage}{0.32\linewidth}
\includegraphics[width=\linewidth]{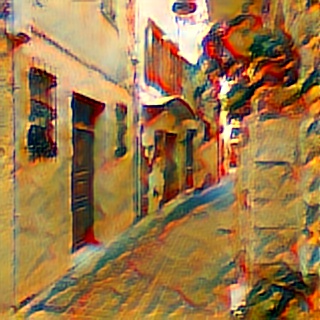}
\end{minipage}
\\
\begin{minipage}{0.32\linewidth}
\includegraphics[width=\linewidth]{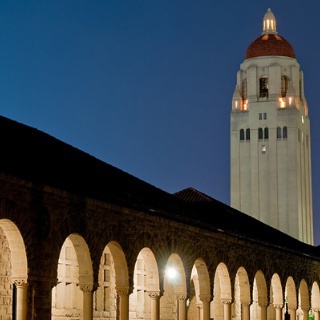}
\end{minipage}
\begin{minipage}{0.32\linewidth}
\includegraphics[width=\linewidth]{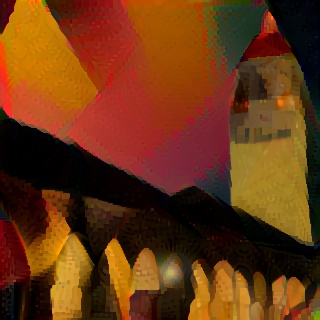}
\end{minipage}
\begin{minipage}{0.32\linewidth}
\includegraphics[width=\linewidth]{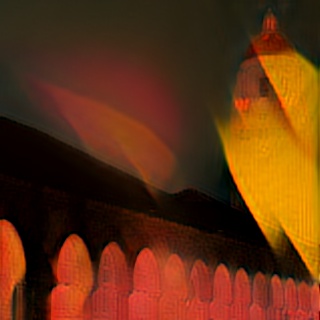}
\end{minipage}
\\
\begin{minipage}{0.32\linewidth}
\includegraphics[width=\linewidth]{inputs/venice}
\end{minipage}
\begin{minipage}{0.32\linewidth}
\includegraphics[width=\linewidth]{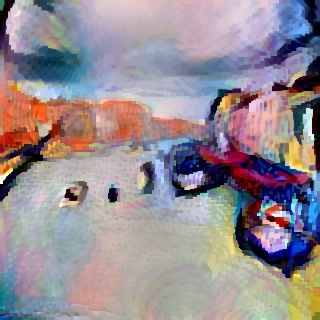}
\end{minipage}
\begin{minipage}{0.32\linewidth}
\includegraphics[width=\linewidth]{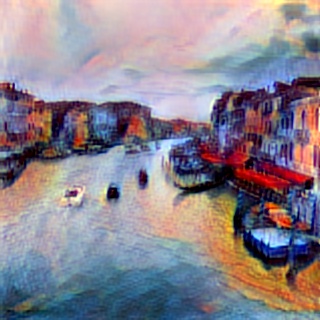}
\end{minipage}
\\
\end{minipage}
\hfill
\vline
\hfill
\begin{minipage}{0.24\linewidth}
\begin{minipage}{\linewidth} 
\centering
\scriptsize
\textbf{Style 1},\\
\emph{Composition VII},\\
W.~Kandinsky, 1913.
\end{minipage}
\\
\begin{minipage}{\linewidth} 
\centering
\includegraphics[width=0.94\linewidth]{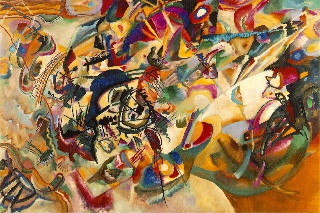}
\end{minipage}
\vspace{3mm}
\\
\begin{minipage}{\linewidth} 
\centering
\scriptsize
\textbf{Style 2},\\
\emph{Self portrait},\\
R.~Magritte, 1923.
\end{minipage}
\\
\begin{minipage}{\linewidth} 
\centering
\includegraphics[height=30mm]{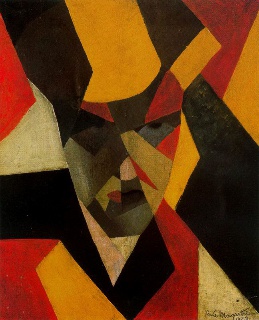}
\end{minipage}
\vspace{3mm}
\\
\begin{minipage}{\linewidth} 
\centering
\scriptsize
\textbf{Style 3},\\
\emph{Woman with a Hat},\\
H.~Matisse, 1905.
\end{minipage}
\\
\begin{minipage}{\linewidth} 
\centering
\includegraphics[height=30mm]{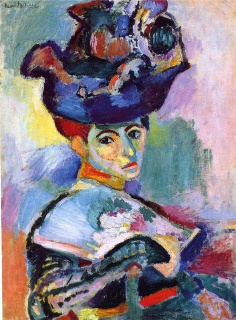}
\end{minipage}
\end{minipage}
\caption{\label{fig:artistic_style_transfer_supp}Artistic style transfer results. First column: Content images. Second: Results obtained with the method of \cite{gatys16}. Third: Our results. Fourth: Style images.}
\end{figure}

We present additional photorealistic style tranfer results in Fig.~\ref{fig:photorealistic_style_transfer_supp} and compare with those obtained in \cite{luan17}. In the first two results in Fig.~\ref{fig:photorealistic_style_transfer_supp}, we notice that the color palette of the results obtained in \cite{luan17} still contains colors present in the original content $\Sig_c$ (blue of the sky in the first result, and yellow of the wall in the second). On the contrary, our results appear more faithful to the color palette of the style images. In the fourth result, we remark that both methods achieve similar qualities. In the last two examples, the color palette of the results obtained in \cite{luan17} seems richer and more faithful to the color palette of the style images than in our results. Nevertheless, the results of \cite{luan17} suffers from artefacts (see, \eg, the sky in the last example). The local affinity constraint imposed via the matting Laplacian $\Lap$ is enforced more strongly in our case thanks to the projection onto ${\rm span}(\Fou_k)$, instead of just a simpler regularisation with $\Lap$ in \cite{luan17}. This explains the reduced number of artefacts but also, sometimes, the limited color palettes in our results.

Finally, we present additional artistic style tranfer results in Fig.~\ref{fig:artistic_style_transfer_supp} to illustrate the fact that the method works on several styles. The results that we obtain are similar to those obtained with the method of \cite{gatys16}.

%
\section{Limitations of the Method}
\label{sec:negative_results}

\begin{figure}
\centering
\begin{minipage}{0.19\linewidth}\centering\scriptsize Content \end{minipage}
\begin{minipage}{0.19\linewidth}\centering\scriptsize \cite{luan17} \end{minipage}
\begin{minipage}{0.19\linewidth}\centering\scriptsize Ours - No filt. \end{minipage}
\begin{minipage}{0.19\linewidth}\centering\scriptsize Ours - Graph filt.\end{minipage}
\begin{minipage}{0.19\linewidth}\centering\scriptsize Style \end{minipage}
\\
\begin{minipage}{0.19\linewidth}
\includegraphics[width=\linewidth]{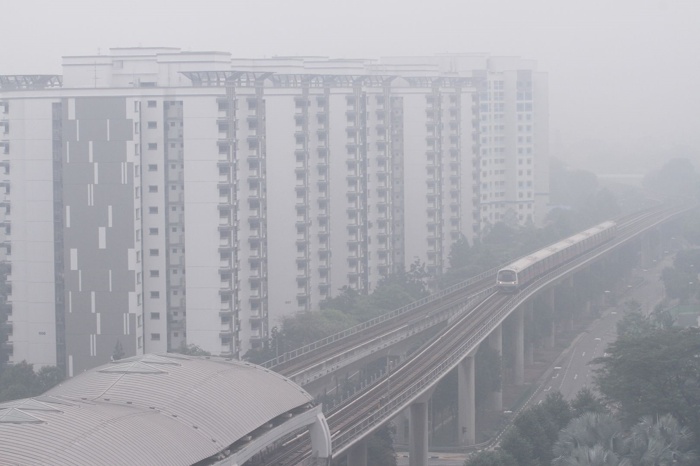}
\end{minipage}
\begin{minipage}{0.19\linewidth}
\includegraphics[width=\linewidth]{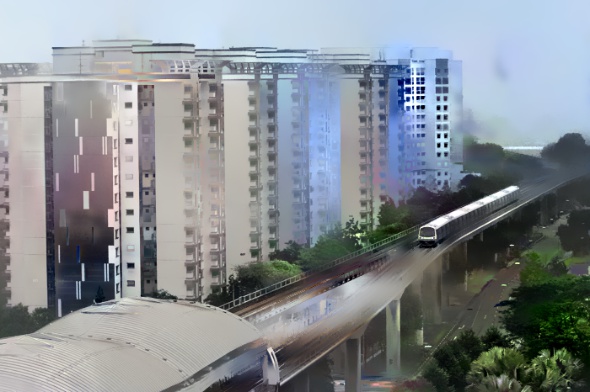}
\end{minipage}
\begin{minipage}{0.19\linewidth}
\includegraphics[width=\linewidth]{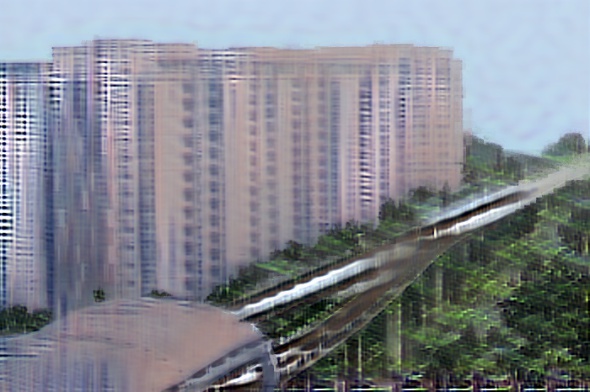}
\end{minipage}
\begin{minipage}{0.19\linewidth}
\includegraphics[width=\linewidth]{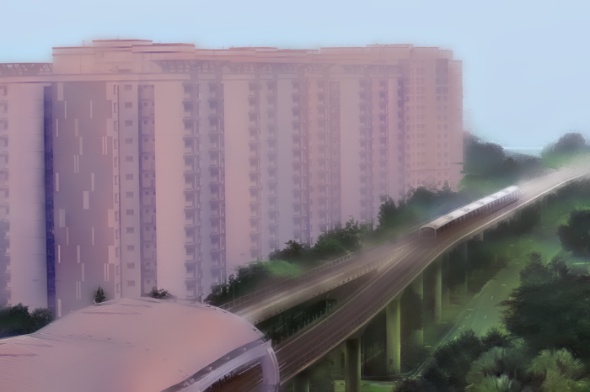}
\end{minipage}
\begin{minipage}{0.19\linewidth}
\includegraphics[width=\linewidth]{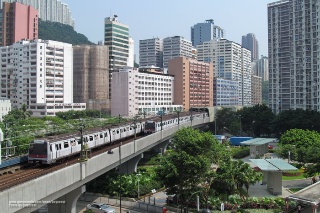}
\end{minipage}
\\
\begin{minipage}{0.19\linewidth}
\includegraphics[width=\linewidth]{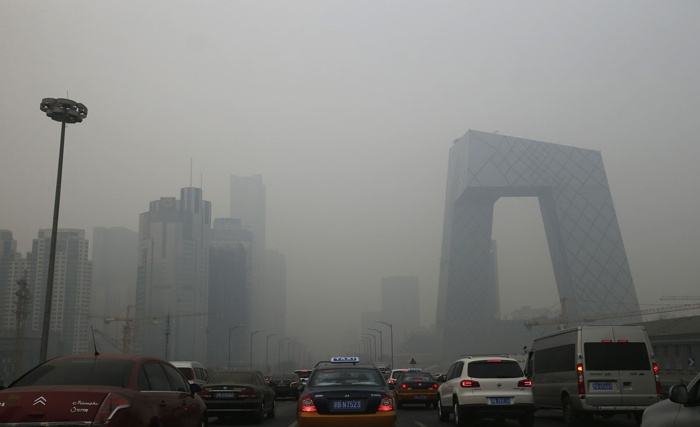}
\end{minipage}
\begin{minipage}{0.19\linewidth}
\includegraphics[width=\linewidth]{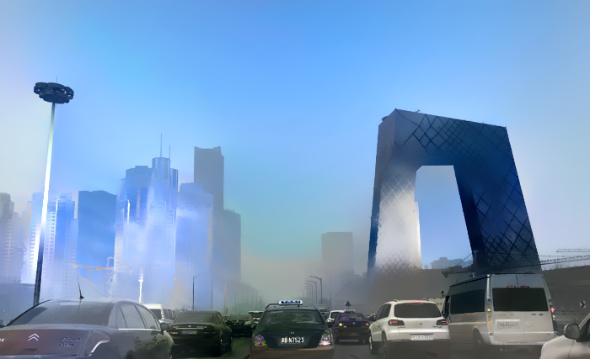}
\end{minipage}
\begin{minipage}{0.19\linewidth}
\includegraphics[width=\linewidth]{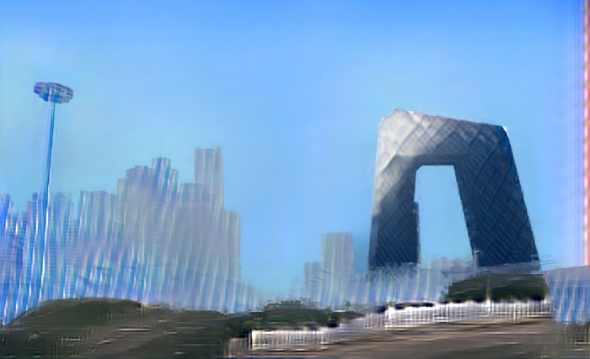}
\end{minipage}
\begin{minipage}{0.19\linewidth}
\includegraphics[width=\linewidth]{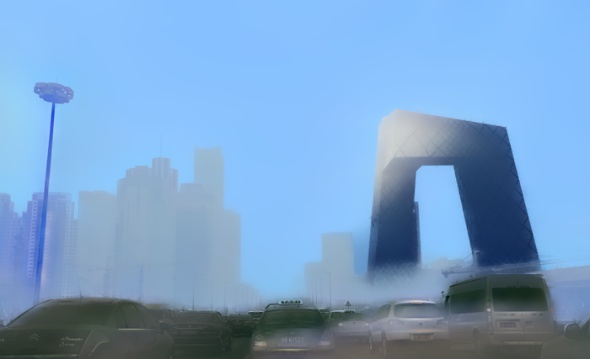}
\end{minipage}
\begin{minipage}{0.19\linewidth}
\includegraphics[width=\linewidth]{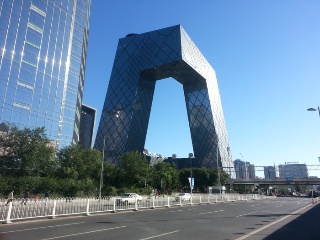}
\end{minipage}
\\
\begin{minipage}{0.19\linewidth}
\includegraphics[width=\linewidth]{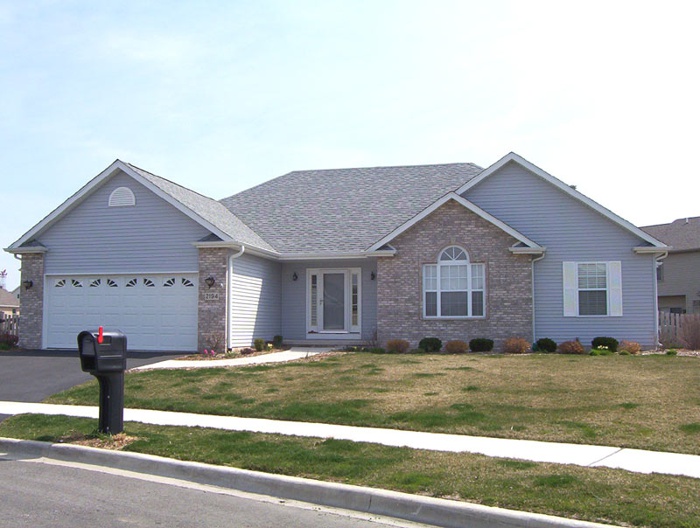}
\end{minipage}
\begin{minipage}{0.19\linewidth}
\includegraphics[width=\linewidth]{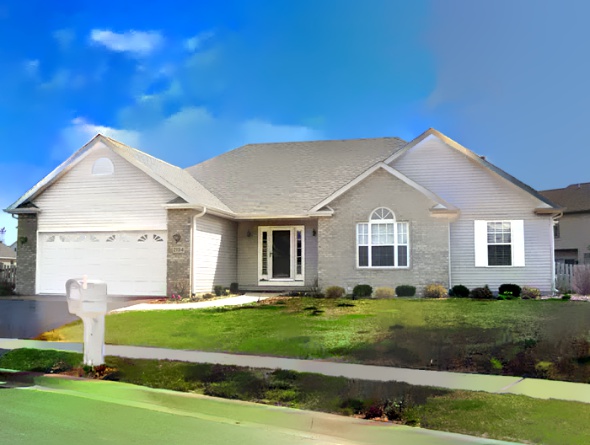}
\end{minipage}
\begin{minipage}{0.19\linewidth}
\includegraphics[width=\linewidth]{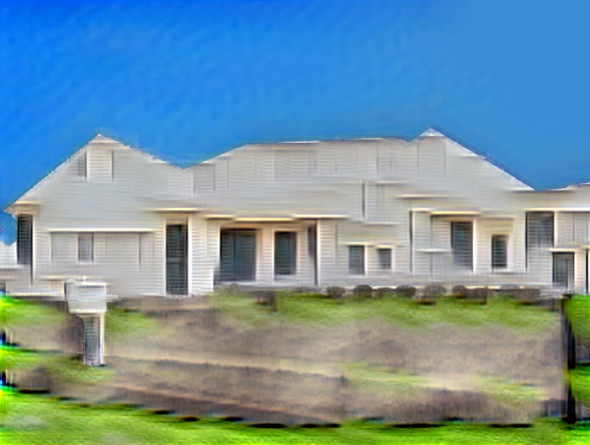}
\end{minipage}
\begin{minipage}{0.19\linewidth}
\includegraphics[width=\linewidth]{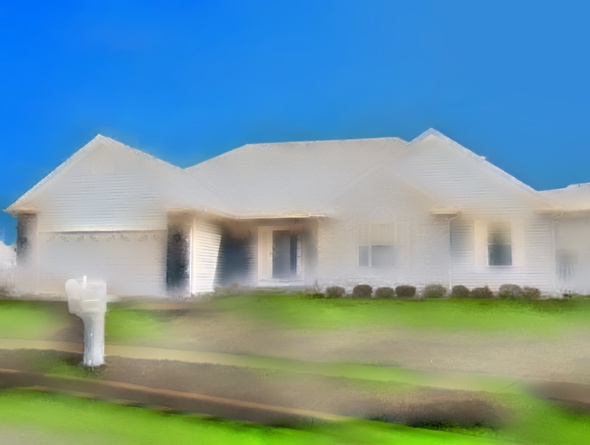}
\end{minipage}
\begin{minipage}{0.19\linewidth}
\includegraphics[width=\linewidth]{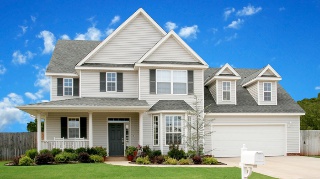}
\end{minipage}
\\
\begin{minipage}{0.19\linewidth}
\includegraphics[width=\linewidth]{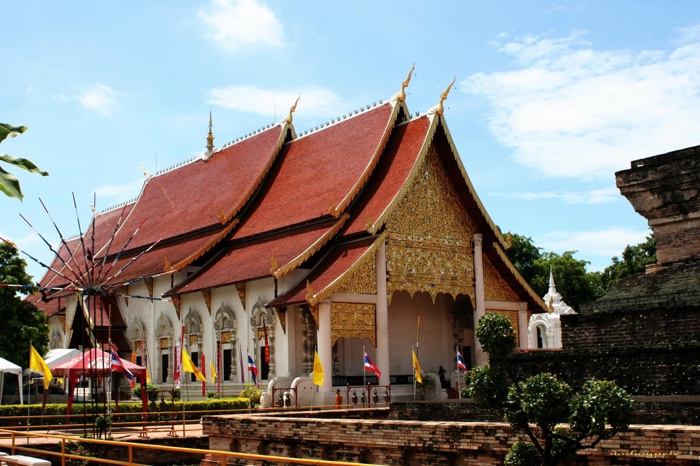}
\end{minipage}
\begin{minipage}{0.19\linewidth}
\includegraphics[width=\linewidth]{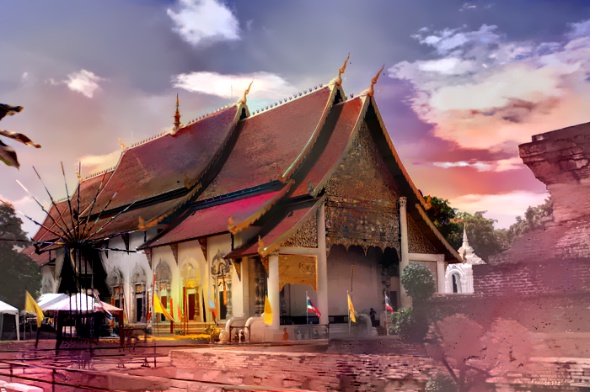}
\end{minipage}
\begin{minipage}{0.19\linewidth}
\includegraphics[width=\linewidth]{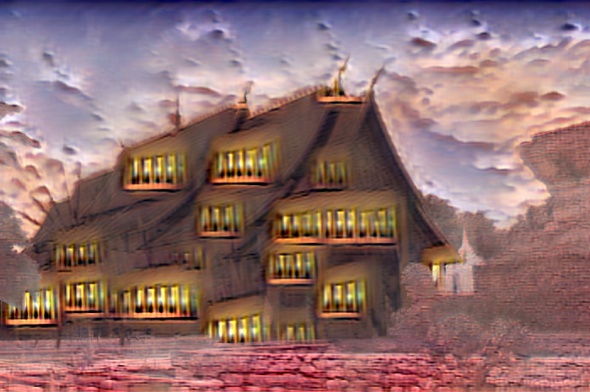}
\end{minipage}
\begin{minipage}{0.19\linewidth}
\includegraphics[width=\linewidth]{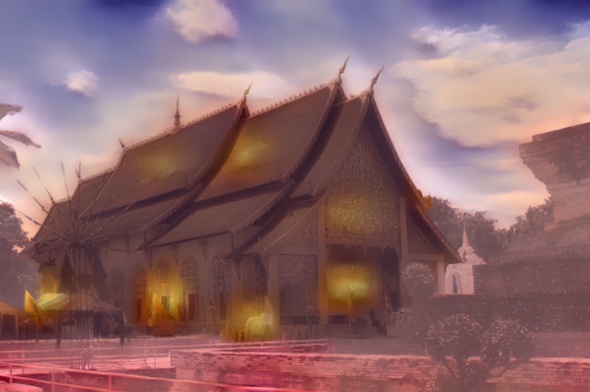}
\end{minipage}
\begin{minipage}{0.19\linewidth}
\includegraphics[width=\linewidth]{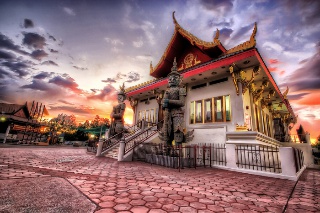}
\end{minipage}
\\
\caption{\label{fig:negative_results}Examples of results where style transfer failed. First column: Content images. Second: Results obtained by \cite{luan17}. Third: Our results \emph{without} graph filtering at runtime. Fourth: Our results \emph{with} graph filtering at runtime. Fifth: Style images. Please refer to Appendix~\ref{sec:negative_results} for comments.}
\end{figure}

In this section, we discuss some style transfer results which are not satisfying. Examples of such results are presented in Fig.~\ref{fig:negative_results}

A typical case where the method fails is when we start from a content image essentially monochromatic and use a colorful style image (see the first two examples in Fig.~\ref{fig:negative_results}). We notice that the color in our results lacks saturation unlike the results in \cite{luan17} (even though we do not judge these results really photorealistic). One of the reasons explaining this lack of saturation is the presence of graph filtering itself. Indeed, the matting Laplacian enforces the transformation to be locally affine; the content image being essentially monochromatic, our graph filtering thus essentially averages locally all the colors. In several examples using $\alpha>1$ permits us to attenuate this effect. Nevertheless, in the first two examples in Fig.~\ref{fig:negative_results}, this lack of color saturation is also visible in the absence of graph filtering. The reason for this behaviour needs further investigation. Some hypotheses are: Not enough monochromatic images in the training datasets; Not enough iteration layers in our network; Not enough feature channels in the forward maps.

Finally, in some cases, the network stylises too differently parts of the content image that were initially similar. This is for example quite noticeable on the houses of the two last results in Fig.~\ref{fig:negative_results} where extra windows appear in the stylised images obtained without graph filtering. The presence of graph filtering in the network in not enough to correct this effect. This filtering averages the style across the image, giving the impression that a veil was added in front of the content image. The reason for this effect was not clearly identified. A likely reason is that the style weight $\lambda_s$ in the training loss was too high for these styles. We noticed that modifying $\alpha$ in \eqref{eq:varying_intensity} at runtime is not sufficient to correct this hypothetical poor choice of $\lambda_s$.

\end{document}